\documentclass{article}

\usepackage[preprint]{neurips_2025}

\usepackage[utf8]{inputenc} 
\usepackage[T1]{fontenc}    
\usepackage{hyperref}       
\usepackage{url}            
\usepackage{booktabs}       
\usepackage{amsfonts}       
\usepackage{nicefrac}       
\usepackage{microtype}      
\usepackage{xcolor}         

\usepackage{bm}
\usepackage{makecell}
\usepackage{longtable}
\usepackage{graphicx}
\usepackage{subcaption}

\usepackage{amsmath}

\newcommand{\comment}[1]{}

\PassOptionsToPackage{numbers, compress}{natbib}

\title{LaDCast: A Latent Diffusion Model for Medium-Range Ensemble Weather Forecasting}

\author{%
  Yilin Zhuang\\
	Department of Aerospace Engineering\\
	University of Michigan\\
	Ann Arbor, MI 48105 \\
	\texttt{ylzhuang@umich.edu} \\
	  \And
    Karthik Duraisamy\\
	Department of Aerospace Engineering\\
	University of Michigan\\
	Ann Arbor, MI 48105 \\
	\texttt{kdur@umich.edu} \\
}

\begin{document}

\maketitle

\begin{abstract}
Accurate probabilistic weather forecasting demands both high accuracy and efficient uncertainty quantification, challenges that overburden both ensemble numerical weather prediction (NWP) and recent machine‑learning methods. We introduce LaDCast, the first global latent‑diffusion framework for medium‑range ensemble forecasting, which generates hourly ensemble forecasts entirely in a learned latent space. An autoencoder compresses high‑dimensional ERA5 reanalysis fields into a compact representation, and a  transformer-based diffusion model produces sequential latent updates with arbitrary‑hour initialization. The model incorporates Geometric Rotary Position Embedding (GeoRoPE) to account for the Earth's spherical geometry, a dual-stream attention mechanism for efficient conditioning, and sinusoidal temporal embeddings to capture seasonal patterns. LaDCast achieves deterministic and probabilistic skill close to that of the European Centre for Medium-Range Forecast IFS‑ENS, without any explicit perturbations. Notably, LaDCast demonstrates superior performance in tracking rare extreme events such as cyclones, capturing their trajectories more accurately than established models. By operating in latent space, LaDCast reduces storage and compute by orders of magnitude, demonstrating a practical path toward forecasting at kilometer‑scale resolution in real time. We open-source our code and models and provide the training and evaluation pipelines at: \href{https://github.com/tonyzyl/ladcast}{https://github.com/tonyzyl/ladcast}.
\end{abstract}

\section{Introduction}

Accurate weather forecasting is critical for various industries and decision-making across sectors. Traditional numerical weather prediction (NWP) models solve the partial differential equations that govern atmospheric fluid dynamics and thermodynamics, achieving high accuracy at the cost of intensive computation and expert‑tuned parameterizations of sub‑grid processes~\cite{stensrud2007parameterization}. These models typically yield a single deterministic trajectory; ensembles are generated by perturbing initial conditions or sampling from observation and parameterization uncertainty~\cite{richardson2024jumpiness}. Furthermore, improving model fidelity requires manual refinement of equations and parameterizations, with little ability to leverage the ever‑growing volume of observational data.

\begin{figure}[htb]
  \centering
  \includegraphics[width=1.0\textwidth]{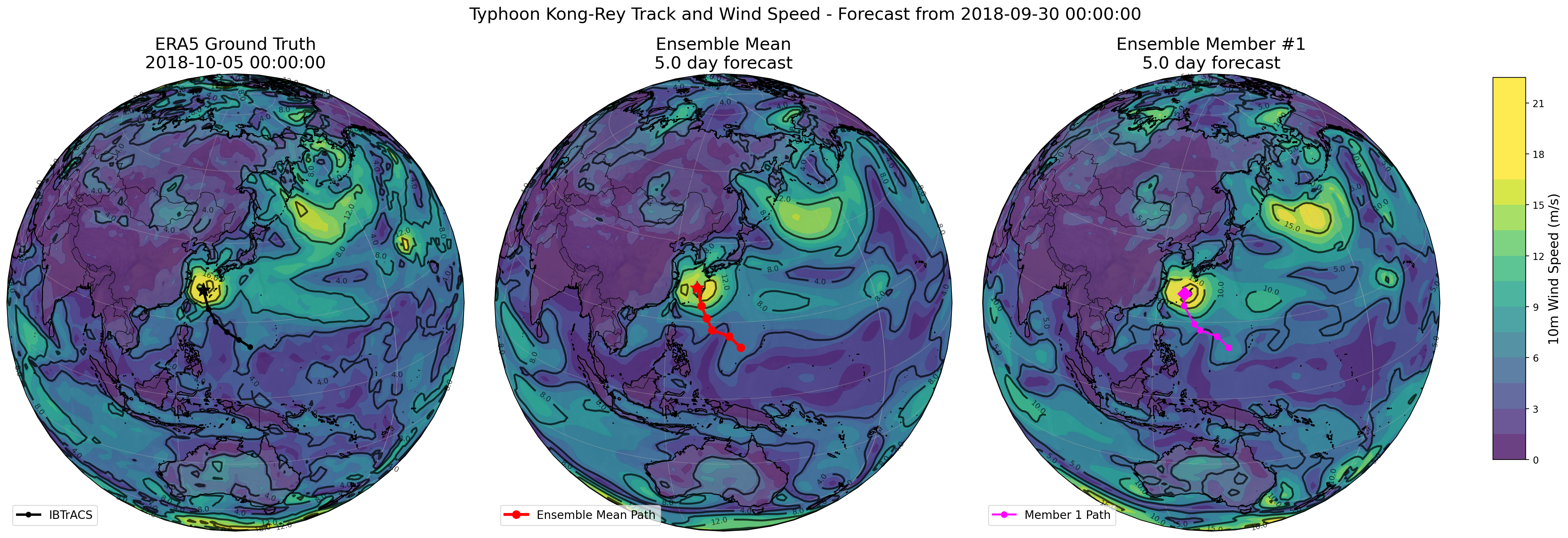}
  \caption{Visualization of a 5-day forecast showing the 10m wind speed forecast for the typhoon Kong-rey. IBTrACS~\cite{knapp2010international} represents the ground truth.}
    \label{fig:globe_vis}
\end{figure}

\begin{figure}[htb]
  \centering
  \includegraphics[width=1.0\textwidth]{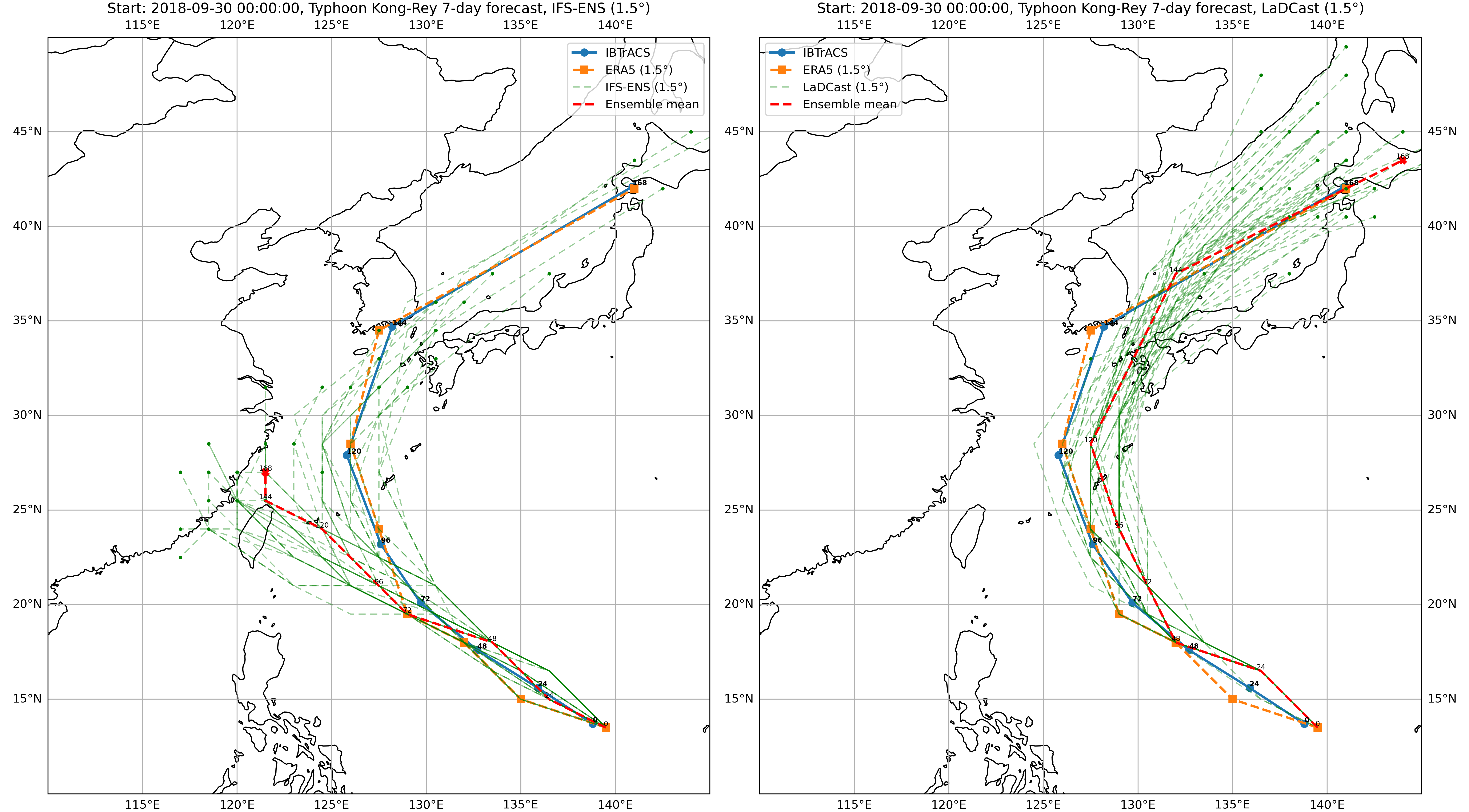}
  \caption{Fifty 10-day forecast trajectories from IFS-ENS and LaDCast for Typhoon Kong-rey (2018), starting from 2018-09-30T00. The IFS-ENS trajectories deviate significantly, with only a few aligning with the true path. In contrast, LaDCast more accurately captures the typhoon’s trajectory.}
    \label{fig:kongrey}
\end{figure}

\vspace{-0.15cm}

Because of the chaotic nature of the atmosphere, predictability is limited to about two weeks~\cite{lorenz1969predictability, judt2018insights}, and medium‑range forecasting models typically provide up to 15‑day predictions. Recently, Machine‑learning-based weather prediction (MLWP) methods~\cite{bi2022pangu, lam2022graphcast, price2023gencast, kochkov2024neural, han2024fengwu, bodnar2024aurora, pathak2022fourcastnet} have emerged as low‑cost alternatives, since long‑term numerical stability is less critical in this context~\cite{ben2024rise}. Most MLWP models train on the highest‐resolution (\(0.25^\circ, 721\times1440\) grids) of the ERA5 reanalysis dataset~\cite{hersbach2020era5}. Transformer‐based architectures have been fine‑tuned on the European Centre for Medium-Range Forecast (ECMWF)’s IFS‑HRES to support even higher resolutions~\cite{han2024fengwu, bodnar2024aurora, malardel2016new}. However, these state‑of‑the‑art models require massive storage (hundreds of terabytes) and often cover only limited time spans, while real-world applications, for instance, wind farm management, demand subkilometer scale resolution with real-time updates. This gap highlights the need for more efficient and scalable approaches, particularly for transformer‑based MLWP, since both resolution and variable count have so far constrained model scaling~\cite{nguyen2024scaling, bodnar2024aurora}, and current models are orders of magnitude smaller than those used in video generation~\cite{kong2024hunyuanvideo}.
While general weather pattern forecasting has seen substantial recent improvements, predicting the trajectory and intensity of rare extreme events such as hurricanes remains a critical challenge with enormous societal impact. Current operational NWP ensemble systems often show significant spread in tracking such events, leading to planning uncertainties for emergency management. An efficient probabilistic forecasting system that can accurately represent uncertainty in extreme weather event trajectories would represent a significant advance for disaster preparedness and mitigation efforts.

A crucial shortcoming remains in producing forecasts that accurately capture uncertainty. The state‑of‑the‑art (SOTA) probabilistic NWP model, ECMWF’s IFS‑ENS~\cite{81142}, initializes an ensemble by perturbing initial conditions and model physics, with each trajectory representing a possible future atmospheric state. Nevertheless, uncertainty quantification is nontrivial and biases can be easily introduced~\cite{81142}. The recent graph‑based diffusion approach, GenCast~\cite{price2023gencast}, is the first MLWP model to surpass IFS‑ENS in probabilistic skill, but it remains expensive to train, uses only a subset of ERA5, and does not support hourly updates.

In this work, we propose a \textbf{la}tent \textbf{d}iffusion framework for probabilistic weather fore\textbf{cast}ing (LaDCast). LaDCast generates ensemble forecasts in a learned latent space, with arbitrary‑hour initialization. It consists of an autoencoder that maps between physical and latent representations and a dual-stream transformer diffusion model~\cite{flux2024, kong2024hunyuanvideo} that generates sequential latent forecasts.

\vspace{-0.1cm}
\noindent {\bf Contributions.} 
LaDCast is, to our knowledge, the first latent-diffusion framework for global weather forecasting, matching SOTA NWP and MLWP models at extended lead times. We introduce Geometric Rotary Position Embedding (GeoRoPE), which adapts rotary embeddings to spherical data by handling longitude periodicity and latitude-dependent feature importance and we apply a dual-stream transformer that processes conditional (initial) and target tokens separately before merging to improve conditioning efficiency. This leads to superior rare event tracking capabilities including exceptional performance in tracking cyclones, producing more accurate trajectory ensembles for typhoons and hurricanes compared to leading NWP systems, offering potential improvements for disaster preparedness applications. Additionally, by performing sequential prediction in latent space, LaDCast’s training cost is orders of magnitude lower than that of NWP and other MLWP models. Finally, we demonstrate that LaDCast’s ensemble distributions match those of top probabilistic NWP systems even without perturbing the initial conditions.

\vspace{-0.5cm}
\section{Related work}
\vspace{-0.5cm}

\noindent {\bf Probabilistic weather forecasting.} Both deterministic NWP and MLWP models produce forecast ensembles by perturbing initial conditions or model physics. For example, ECMWF’s IFS‑ENS perturbs the initial state and physical parameterizations to generate an ensemble, while GenCast~\cite{price2023gencast} explores heuristic Gaussian‑noise perturbations and demonstrates improved skill at longer lead times. IFS‑ENS remains computationally expensive and issues forecasts only every twelve hours, and although full‑field MLWP models have low inference cost, their high training expense limits community access. Building on the one‑to‑one predictions of Pangu‑Weather~\cite{bi2022pangu}, recent MLWP methods have shifted to two‑to‑one sequential architectures~\cite{price2023gencast, lam2022graphcast, bodnar2024aurora}, performing iterative rollouts for extended‑range forecasts. Despite these advances, seq‑to‑seq models remain underexplored in weather forecasting,  though they have shown strong performance in general time‑series tasks~\cite{zeng2023transformers}.

\noindent {\bf Weather data and neural compression.} Traditional weather data compression techniques achieve substantial size reduction via linear quantization of 64‑bit floats~\cite{WMO2003GRIB2}, but offer limited benefit for MLWP pipelines. Although most full‑field MLWP models employ skip and residual connections to preserve high‑frequency features~\cite{bi2022pangu, han2024fengwu, bodnar2024aurora}, they still rely on separate compression–decompression modules to manage high‑dimensional inputs. Neural compressors like CRA5~\cite{han2024cra5} have shown promise: by encoding ERA5 into a latent space based on a variational autoencoder (VAE)~\cite{kingma2013auto}, autoregressive forecasts match IFS‑ENS performance, yet CRA5’s spatial compression remains modest and its VAE architecture can limit probabilistic forecasting. More recent work~\cite{chen2024deep, xie2024sana} demonstrates that autoencoder latent spaces are also suited for latent diffusion models~\cite{rombach2022highresolutionimagesynthesislatent}. In particular, the Deep Compression Autoencoder (DC‑AE)~\cite{chen2024deep} outperforms the standard SD‑VAE in spatial compression, thereby reducing downstream computational costs. To date, however, such neural‑compression strategies have not been applied to global weather‑forecasting systems.

\noindent {\bf Diffusion models for weather forecasting.} Diffusion models excel at sampling from complex distributions~\cite{ho2020denoising, song2021scorebased}, and latent diffusion models (LDMs) scale efficiently to high‑resolution data by operating in a compressed latent space~\cite{rombach2022highresolutionimagesynthesislatent}. This reduction in computational cost makes LDMs particularly attractive for building higher‑resolution forecasting systems. In weather contexts, GenCast~\cite{price2023gencast} represents the current state of the art, surpassing the IFS-ENS, using a full‑space diffusion model with two‑to‑one sequential rollouts. Recently, ArchesWeather~\cite{couairon2024archesweather} builds a diffusion model on top of deterministic model outputs and have outperformed IFS-ENS with the drawback on overfitting of the deterministic model. Other full‑field diffusion efforts address precipitation nowcasting~\cite{yu2024diffcast} and global climate emulation~\cite{ruhling2023dyffusion, ruhling2024probablistic}. To further curb computation costs, LDMs have also been explored for ERA5 data assimilation~\cite{andry2025appa} and precipitation nowcasting~\cite{leinonen2023latent, gao2023prediff}. Recent work applies LDM‑based autoregressive solvers to partial differential equations~\cite{jacobsen2023cocogenphysicallyconsistentconditionedscorebased, gao2025generative, huang2024diffusionpdegenerativepdesolvingpartial, du2024conditional}, yet no existing study leverages LDMs for global, multi‑variable, multi‑pressure‑level weather forecasting.

\section{Background and preliminaries}

\noindent {\bf Problem setup.}  We consider probabilistic spatiotemporal forecasting over a sequence of $T$ high‐dimensional fields \(\{x^t\}_{t=1}^T\subset\mathcal{X}\), where \(x^t\in\mathbb{R}^{C\times H\times W}\) is the space representing the weather data at time \(t\), with \(C\) channels, \(H\) height (latitude), and \(W\) width (longitude). To work in a compact latent space, we employ an autoencoder with encoder \(\phi_E:\mathcal{X}\to\mathcal{Z}\) and decoder \(\phi_D:\mathcal{Z}\to\mathcal{X}\), such that each observation is mapped as \(z^t=\phi_E(x^t)\in\mathcal{Z}\) and approximately reconstructed via \(\phi_D(z^t)\). We estimate the conditional distribution of the  \(h\) latent variables given the previous \(\ell\) snapshots:
\begin{equation}
p\bigl(z^{t+1:t+h}\,\bigm|\,z^{t-\ell+1:t}\bigr)
\;=\;
\prod_{i=1}^{h}
p\,\!\bigl(z^{t+i}\,\bigm|\,z^{t-\ell+i:t+i-1}\bigr).
\end{equation}

\noindent {\bf Latent compression.} To reduce the dimensionality of the weather fields while retaining the essential spatial structure, we train a DC-AE~\cite{chen2024deep} with a relative loss \(L_2\). We treat each snapshot as an independent sample for arbitrary sequence selection, the training objective can be formulated as:
\[
\min_{\theta_E,\theta_D}\;\mathbb{E}_{x\sim p_{\mathrm{data}}}\;\mathcal{L}\bigl(\phi_D(\phi_E(x)),\,x\bigr),\,\, \text{with} \,\, \mathcal{L}(x,\hat x)
=\frac{\|x-\hat x\|_{2}}{\|x\|_{2}}.
\]
The DC-AE consists of convolutional and transformer layers. We modify the convolutional kernels to account for the spherical nature of the data, as shown in Figure~\ref{fig:pole}. Static channels that derived from the Earth's surface are convolved with variable channels, which can be regarded as adding a static embedding to the variable channels at the first layer (see Appendix~\ref{app:autoencoder}).

\noindent{\bf Conditional Diffusion model.} The diffusion model consists of forward and reverse processes. The forward process applies Gaussian noise to the data, gradually transforming it into a Gaussian process. The reverse process learns to denoise the data step by step, ultimately generating new samples within the data distribution. It has been shown that the variance-exploding formulation~\cite{song2021scorebased} is  suitable for handling cases~\cite{zhuang2025spatially} where  physical fields could have large variance despite being normalized. A variation of the variance-exploding formulation is proposed~\cite{karras2022elucidating} to train a denoising function, \(D(\bm{z}^{out}; \bm{z}^{in},\sigma)\), where \(\sigma\) is the noise level and we denote \(z^{t-\ell+i:t+i-1}\) as \(\bm{z}^{in}\), and \(z^{t+1:t+h}\) as \(\bm{z}^{out}\). The \(L_2\) denoising error can be formulated as:
\begin{equation}
    \mathbb{E}_{\bm{z}^{out}_0 \sim p_{\text{data}}} \mathbb{E}_{\bm{n} \sim \mathcal{N} (0,\sigma^2 I)} \|D(\bm{z}^{out}_0 + \bm{n}; \bm{z}^{in},\sigma) - \bm{x}_0\|^2_2,
\end{equation}
where \(p_{\text{data}}\) denotes the data distribution and \(\bm{n}\) is the added noise. The denoising function \(D\) is parameterized by a neural network, \(D_\theta\), and the model is trained to minimize the denoising error at noise level \(\sigma\) on the latent sequence, \(\bm{z}^{out}\), with conditioning on the input sequence, \(\bm{z}^{in}\).

We follow the preconditioning introduced in EDM~\cite{karras2022elucidating} to scale the denoising function.  At noise level \(\sigma\), the denoiser \(D_\theta\) acting on the output block \(\bm{z}^{out}\) with conditioning on the input block \(\bm{z}^{in}\) is written as
\begin{equation}
  D_\theta\bigl(\bm{z}^{out};\,\bm{z}^{in},\sigma \bigr)
  = c_{\mathrm{skip}}(\sigma)\,\bm{z}^{out}
  + c_{\mathrm{out}}(\sigma)\,
    F_\theta\!\bigl(
      c_{\mathrm{in}}(\sigma)\,\bm{z}^{out};\,
      \bm{z}^{in},\,
      c_{\mathrm{noise}}(\sigma)
    \bigr),
\end{equation}
where \(F_\theta\) is a neural network and \(c_{\mathrm{skip}}(\sigma)\), \(c_{\mathrm{in}}(\sigma)\), \(c_{\mathrm{out}}(\sigma)\) and \(c_{\mathrm{noise}}(\sigma)\) are predefined, scaling factors of \(\sigma\) that rescale the neural network’s input, output, and skip connection when denoising \(\bm{z}^{out}\).

Rather than sampling from the corresponding stochastic differential equation, we use the probability‐flow ODE formulation~\cite{song2021scorebased}, which yields a deterministic sampler. The forward process of the EDM is defined as $
  \bm{z}^{out} = \bm{z}^{out}_0+\sigma_t\bm{n},$
with the probability‐flow ODE
\begin{equation}
\label{eq:pf_ode}
  d\bm{z}^{out}_{-} 
  = -t \,\nabla_{\bm{z}^{out}}\log p_t(\bm{z}^{out}\!\mid\!\bm{z}^{in}, \sigma)\,dt = \frac{\bm{z}^{out} - D_\theta(\bm{z}^{out};\bm{z}^{in},\sigma)}{\sigma^2}\,dt.
\end{equation}


\section{LaDCast: Latent Diffusion for Weather Forecasting}

We select six single variables, six atmospheric variables and five static features from the ERA5 reanalysis dataset~\cite{hersbach2020era5} and train our model on \(1.5^\circ \, (121\times240 \text{grids})\) to build our model. The six single variables are 10-meter U and V components of wind (10u, 10v), 2-meter temperature (2t), mean sea level pressure (msl), sea surface temperature (sst) and 6-hour total precipitation (tp-6hr, derived from the total precipitation). The six atmospheric variables are geopotential (z), specific humidity (q), temperature (t), U and V components of wind (u, v) and vertical velocity (w), each at 13 pressure levels (1000, 925, 850, 700, 500, 400, 300, 250, 200, 150, 100, 70, 50 hPa). The five static features are land–sea mask (lsm), standard deviation of orography (sdor), angle of sub-gridscale orography (isor), anisotropy of sub-gridscale orography (anor) and slope of sub-gridscale orography (slor). A summary of these features is provided in Table~\ref{tab:variables}.

The DC-AE has 256M parameters and it takes 40 H100 days\footnote{Here we provide the reproducible estimated training time based on a separate run with different number of GPUs, as our system is capped by data throughput.} to train from 1979 to 2017 and 9 H100 days to finetune the decoder on the same period. The largest LaDCast model that we trained has 1.6B parameters, despite its size, it takes only 5.3 H100 days to train the diffusion model on the same period on the reduced latent space. In comparison, Stormer~\cite{nguyen2024scaling} takes 64 H100 days to train on \(1.4^\circ\) and ArchesWeatherGen~\cite{couairon2024archesweather} takes 10.3 H100 days to train on \(1.5^\circ\) data, following the conversion between different GPUs used in~\cite{couairon2024archesweather}. For the 1.6B LaDCast model, a 15-day forecast takes under one minute to generate on an H100 and since the forecast is performed in the latent space, multiple ensembles can be rolled out in parallel on a single GPU.

The time interval between consecutive output sequences is 6 hours, and the model is trained hourly to initialize at arbitrary times. The default configuration uses a one-to-four sequence, and we do not employ multistep finetuning~\cite{lam2022graphcast} during training. The model is rolled out 15 times to generate 15-day ensemble forecasts. 

\vspace{-0.15cm}
\subsection{Model architecture}
\vspace{-0.15cm}

\begin{figure}[htb]
  \centering
  \includegraphics[width=1.0\textwidth]{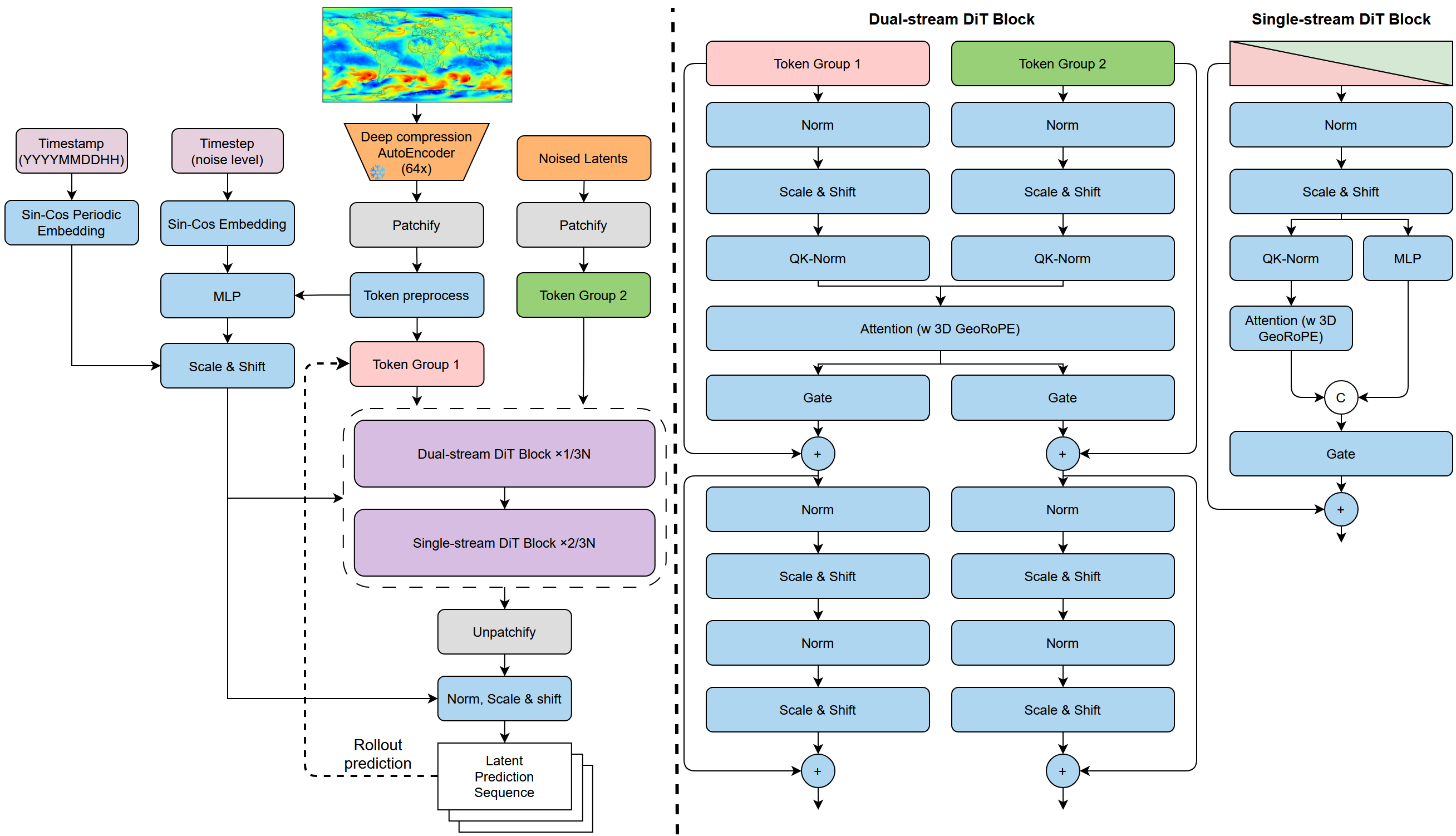}
  \caption{Architecture of the LaDCast model: Predictions are made in the latent space and the conditioning tokens are concatenated with the target denoising tokens for joint attention.}
    \label{fig:scheme}
\end{figure}

\noindent {\bf Inputs.} The latent-space sequence has shape \(T\times C\times H\times W\). We apply a 3D convolution to patchify the spatial and temporal dimensions. The patch size for both spatial and temporal axes is set to one, as smaller patches have been shown to improve autoregressive performance~\cite{nguyen2024scaling} resulting in an overall \(T\times H \times W\) patches per token group, depending on sequence length. In addition to the diffusion-process timestep (which corresponds to the noise level), we introduce a sinusoidal periodic embedding for the elapsed year progress as conditioning (see Appendix~\ref{app:arch_ladcast}). 

\noindent {\bf Position embedding.} Since our latent space preserves the spatial structure of the input, we adapt the Rotary Position Embedding (RoPE)~\cite{su2024roformer} to better account for the spherical geometry of the data, producing Geometric RoPE (GeoRoPE). GeoRoPE splits positional encoding into latitude and longitude components: Inspired by heuristic atmospheric-circulation window attention~\cite{han2024cra5}, the latitude embedding emphasizes high‑frequency features, while the longitude embedding uses low‑frequency components to capture periodicity (see Appendix~\ref{app:arch_ladcast} for details). To extend GeoRoPE to 3D, we partition each query and key vector into three segments, treating the temporal index as the third positional component.

\noindent {\bf Conditioning on the initial profile.} The patchified conditioning tokens are first preprocessed using DiT blocks with GeoRoPE and then passed to the transformer backbone along with the timestep and elapsed year progress embeddings. The dual-to-single stream architecture, introduced in~\cite{flux2024} for video generation, is adopted here: the dual-stream DiT block consists of two branches, allowing the model to separately encode conditioning and target tokens before merging them.

\section{Experiments}

\noindent {\bf Dataset.} We train and evaluate LaDCast on the ERA5 reanalysis dataset~\cite{hersbach2020era5} at \(1.5^\circ\) resolution. The \(1.5^\circ\) dataset is hosted by WeatherBench2 (WB2)~\cite{rasp2024weatherbench} and is downsampled from the original \(0.25^\circ\) dataset. During preprocessing, we compute the hourly 6‑hour total precipitation, which shifts the first available training sample to 1979-01-01T05. For full details, see Appendix~\ref{app:data}.

\noindent {\bf Training.} We train two LaDCast models, one with 1.6B parameters and one with 375M parameters. The 375M model is used for the ablation study, while all other experiments (unless otherwise noted) use the 1.6B model. To improve throughput, we train the DC‑AE using data augmentation, with a slight trade‑off in convergence speed. We train the DC-AE for 30 epochs and then fine-tune its decoder for an additional 10 epochs. Next, we train LaDCast for 10 epochs. During training of both models, we apply an exponential moving average to the weights and do not perform model selection based on validation performance. For full details on the training procedure and hyperparameters, see Appendix~\ref{app:train}.

\noindent {\bf Evaluation.} We evaluate the model daily on UTC 00/12 over the year 2018. For the ablation study, we sample ten days per month and evaluate at the same two times. All reported metrics, root mean square error (RMSE) and continuous ranked probability score (CRPS)~\cite{gneiting2007strictly} are latitude-weighted, further information on their computation is given in Appendix~\ref{app:metric}. For ensemble forecasts, we generate 50 members for the main experiments and 20 members for the ablation study.

\noindent {\bf Compressing the ERA5}.
\begin{table}[ht]
  \centering
  \caption{Compression methods comparison.}
  \label{tab:compression}
  \begin{tabular}{@{}l c *{6}{c} c @{}}
    \toprule
    Method/Value & Year
      & \multicolumn{6}{c}{Weighted RMSE}
      & Spatial comp.\ ratio \\
    \cmidrule(lr){3-8}
    & 
      & z500 & t850 & 10v & 10u & 2t & msl
      &  \\
    \midrule
    Mean &  1979-2017 & 5.4E4 & 274 & 0.19 & -0.05 & 278 & 1.0E5 & \\
    \midrule
    VAEformer~\cite{han2024cra5}
      & 2018-
      & 26.15 & 0.58 & 0.45 & 0.49 & 0.74 & 28.02
      & 16
       \\
    \midrule
    DC-AE & 2018 & 28.96 & 0.65 & 0.54 & 0.55 & 0.78 & 30.32 & 64 \\
    \bottomrule
  \end{tabular}
\end{table}

For comparison, VAEformer~\cite{han2024cra5} is trained at a similar resolution of \(1.4^\circ\) with 69 channels. As shown, the DC‑AE achieves comparable performance while delivering a higher spatial compression ratio. The actual compression ratio for both models equal to their spatial compression ratio, with the DC-AE’s latent representation having a shape of \(84\times15\times30\). Table~\ref{tab:rmse_by_year} summarizes the reconstruction loss for all variables from 2018 to 2022. We observe that high‑frequency components, such as velocity and specific humidity, remain stable over this period, whereas correlated variables such as potential and temperature exhibit noticeable degradation. For example, the RMSE for mean sea level pressure increases by 10\% from 2018 to 2022.

\noindent{\bf Evaluation with deterministic metrics.}
\begin{figure}[htb]
  \centering
  \includegraphics[width=1.0\textwidth]{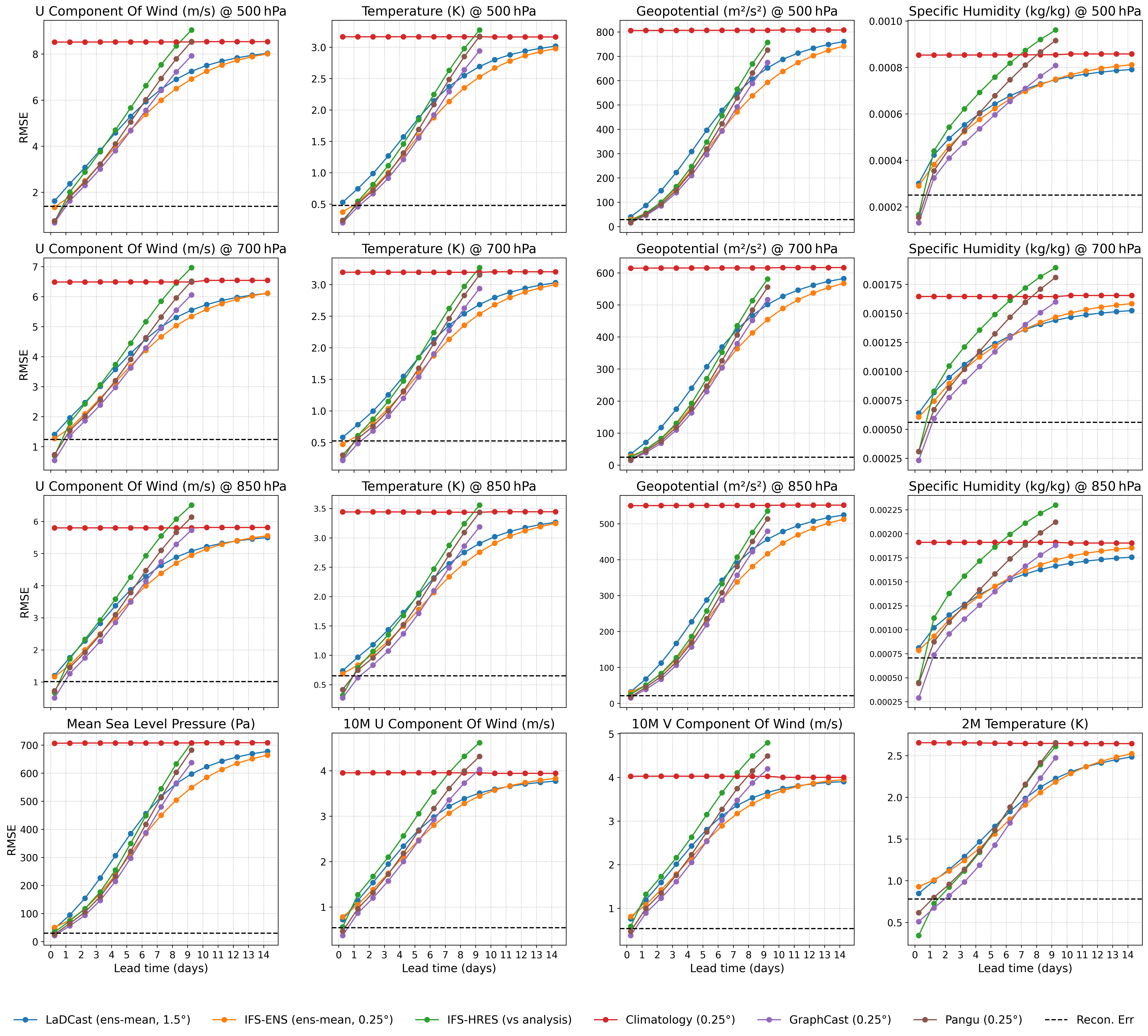}
  \caption{Latitude-weighted RMSE for 2018. The LaDCast shows better performance than deterministic models at longer lead times, the latent compression is a main source of error for LaDCast.}
    \label{fig:2018_rmse}
\end{figure}
In 2018, the only probabilistic model hosted on WB2 is IFS-ENS. Therefore, we compute the ensemble mean of both IFS-ENS and LaDCast for comparison with the remaining deterministic models. All models are evaluated against ERA5, except IFS-HRES, which is compared to its own analysis data for consistency~\cite{lam2022graphcast}. Figure~\ref{fig:2018_rmse} shows that LaDCast and IFS‑ENS outperform the other models at longer lead times and converge toward climatology, i.e., the long‑term mean. Since LaDCast does not use perturbed initial conditions, its initial deviation primarily due to latent compression, whereas for IFS‑ENS it stems from its perturbations. A comparison on the RMSE of year 2018 and 2019 is shown in Figure~\ref{fig:ladcast_2018_2019}, the overall performance remains stable in 2019, with some variables exhibiting lower RMSE than in 2018, demonstrating that the proposed method generalizes well across years.

\begin{figure}[htb]
  \centering
  \includegraphics[width=1.0\textwidth]{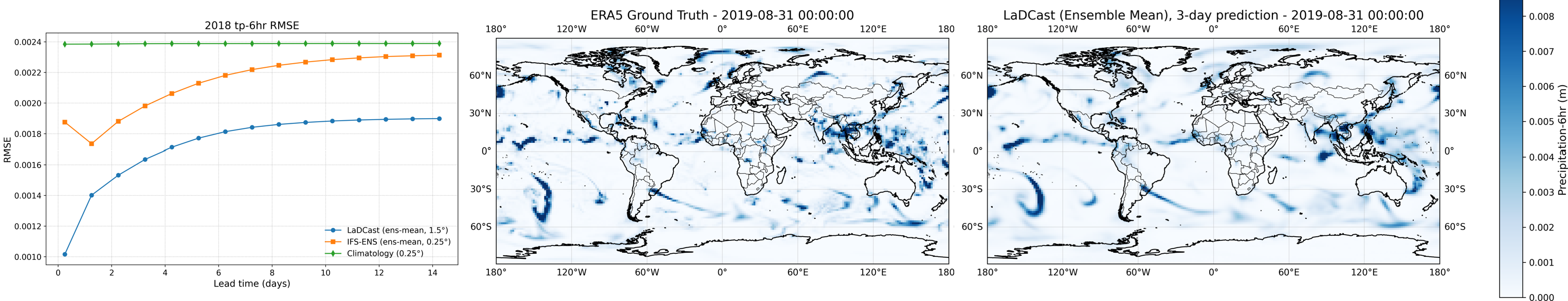}
  \caption{Total precipitation 6hr RMSE for 2018 compared to IFS-ENS. The visualization shows a 3-day forecast during the 2019 Hurricane Dorian.}
    \label{fig:tp_6hr}
\end{figure}

In addition, we found the RMSE of total precipitation (6hr accumulation) performs significantly better than the IFS-ENS, possibly due to the hourly training scheme. The RMSE and a qualitative example are shown in Figure~\ref{fig:tp_6hr}.

\begin{figure}[htb]
  \centering
  \begin{subfigure}[t]{\linewidth}
    \centering
    \includegraphics[width=\linewidth]{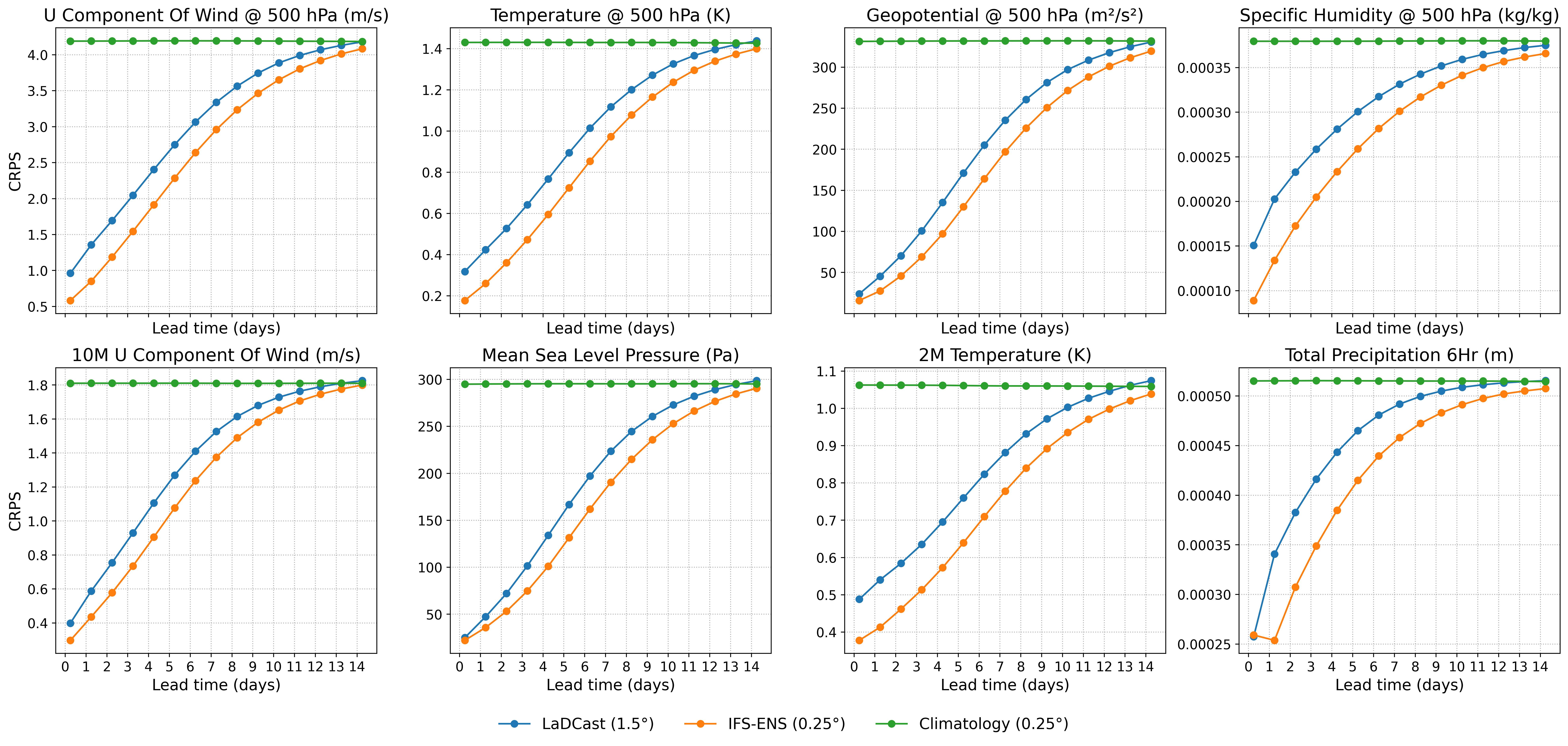}
    \label{fig:2018_crps}
  \end{subfigure}
  \\[-1ex]
  %
  \caption{%
  Results of the CRPS$(\downarrow)$ for 2018, Due to the deterministic conditioning, the spread of LaDCast is lower than IFS-ENS, which results in a higher CRPS value.
  }
  \label{fig:combined_prob_2018}
\end{figure}

\vspace{-0.15cm}

\noindent{\bf Probabilistic forecasting.}
For probabilistic forecasting, we report the CRPS which measures how well the ensemble captures the distribution. The CRPS metric is shown in Figure~\ref{fig:combined_prob_2018}. LaDCast does not arbitrarily perturb the conditioning initial profile, thus has a smaller spread than IFS-ENS. We present three storm tracking case studies: Typhoon Kong-rey (2018), Hurricane Dorian (2019) and Hurricane Lorenzo (2019). The training data covers the period 1979-2017, so all three storms lie outside the training range. Storm center trajectories are extracted using a heuristic tracker based on mean sea level pressure and geopotential height, similar to the method used in~\cite{bodnar2024aurora} based on mean sea level pressure and geopotential (see Appendix~\ref{app:tracking} for details). Tracked hurricane positions are snapped to the nearest grid point, and we overlay these with the ERA5-derived positions for verification. Ground truth tracks are taken from the IBTrACS~\cite{knapp2010international}.
\begin{figure}[htb]
  \centering
  \includegraphics[width=1.0\textwidth]{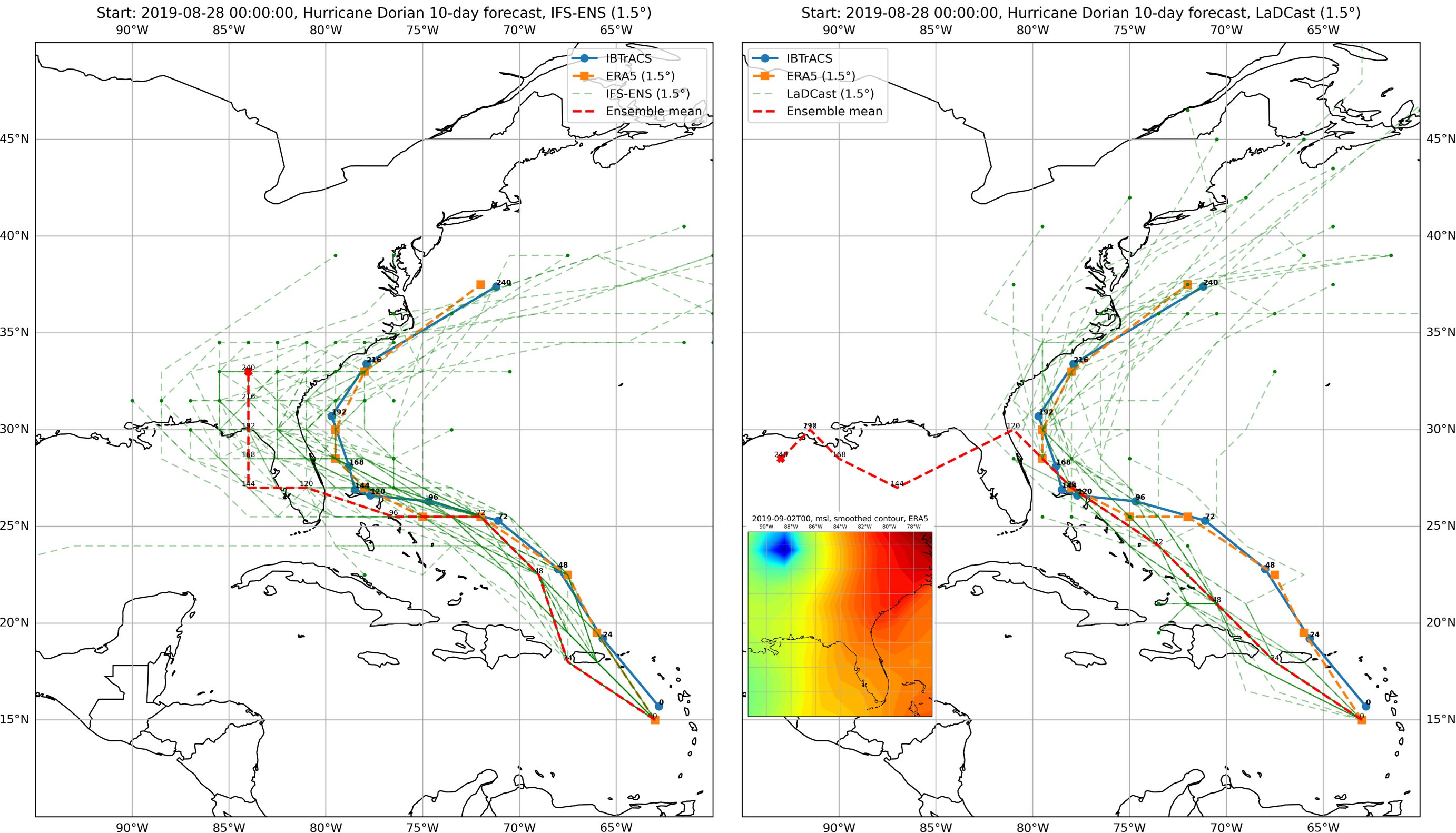}
  \caption{50 trajectories, 10-day forecast of IFS-ENS and LaDCast for hurricane Dorian (2019), starting from 2019-08-28T00. The IFS-ENS shows a large spread of the trajectories, and predicting a majority of the trajectories heading Florida. The deviation in the LaDCast ensemble mean track is due to the heuristic tracker detecting a nearby pressure minimum not located at the actual center.}
    \label{fig:dorian}
\end{figure}

\vspace{-0.15cm}

\begin{figure}[htb]
  \centering
  \includegraphics[width=1.0\textwidth]{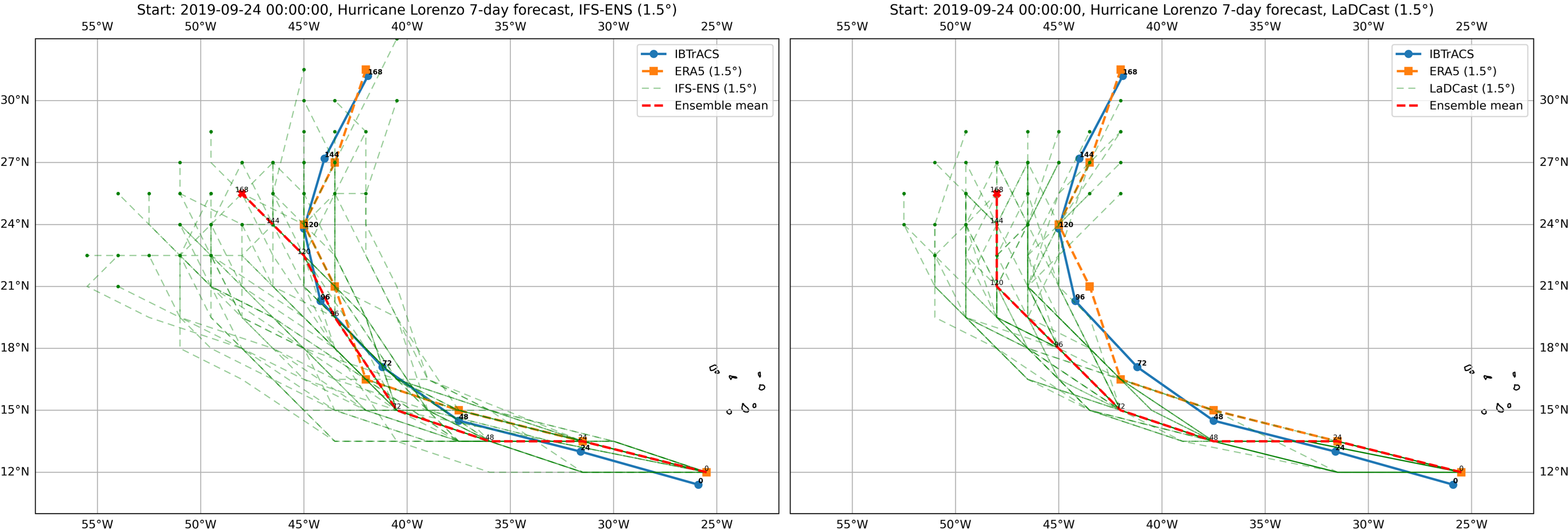}
  \caption{50 trajectories, 7-day forecasts from IFS-ENS and LaDCast for Hurricane Lorenzo (2019), initialized at 2019-09-24T00. Most IFS-ENS trajectories fail to predict the northeastward heading.}
  \label{fig:lorenzo}
\end{figure}

\vspace{-0.15cm}

Figure~\ref{fig:kongrey} shows the 7-day tracking results for Typhoon Kong-rey (2018): the LaDCast ensemble more accurately captures the typhoon’s path than IFS-ENS. A more challenging example is the 10-day forecast of Hurricane Dorian (2019) in Figure~\ref{fig:dorian}: IFS-ENS exhibits a wide spread and directs most trajectories toward Florida, whereas LaDCast follows the true track more closely, sending most members northward along the U.S. East Coast. However, the ensemble-mean forecast erroneously heads westward. Closer inspection reveals that the heuristic tracker, which uses minimum mean sea-level pressure and geopotential, is misled by a low-pressure signal over Florida (see the inset pressure plot in Figure~\ref{fig:dorian}), while the IBTrACS dataset places the storm center offshore. We further verified specific humidity fields, confirming that the extracted LaDCast tracks are correct and that the IFS-ENS extraction aligns with~\cite{price2023gencast}. This case highlights the limitations of relying solely on the ensemble mean in extreme-event forecasting. Ensemble spread is essential for capturing trajectory uncertainty in cyclones. Hurricane Lorenzo (2019) is an extreme out-of-distribution example, as it is the easternmost Category 5 Atlantic storm on record. In the 7-day forecasts shown in Figure~\ref{fig:lorenzo}, IFS-ENS again shows a wide spread, with only a few members tracking northeastward. By contrast, the LaDCast ensemble more accurately reproduces the observed path, despite minor deviation in the ensemble mean.

\noindent{\bf Ablations of LaDCast}: \noindent{\em a) Model scaling and ensemble size:} Figure~\ref{fig:ladcast_scaling} evaluates the effects of model size and ensemble size on LaDCast’s performance. The results show that larger models yield better performance and that increasing the ensemble size provides modest additional gains. This demonstrates the feasibility of applying the LaDCast framework in scenarios with higher grid resolutions and finer timescales. Figure~\ref{fig:ladcast_input_sequence} examines how the length of output sequences affects robustness. The default 1-to-4 configuration outperforms both 1-to-2 and 1-to-8 settings. \noindent{\em b) Ensemble generation:} Figure~\ref{fig:ladcast_noise} examines the effect of perturbing latent initial conditions; such perturbations do not improve overall performance. Figure~\ref{fig:ladcast_steps} shows that increasing the number of reverse-sampling steps enhances the reconstruction of high-frequency components, with 20 steps providing a good balance between quality and computational efficiency. 

\section{Conclusions and future work}

We present LaDCast, the first latent diffusion model for global probabilistic weather forecasting. LaDCast generates ensemble forecasts in a learned latent space with initialization at arbitrary hours. The framework comprises an autoencoder mapping between physical and latent representations, and a dual-stream transformer diffusion model that produces sequential latent forecasts. We demonstrate that LaDCast achieves performance comparable to state-of-the-art NWP models at longer lead times while requiring orders of magnitude less computational cost. Case studies in cyclone tracking show that our unperturbed conditional generation yields more accurate storm tracks than IFS-ENS. Overall, although our experiments here use a relatively coarse \(1.5^\circ\) grid, LaDCast’s latent-space framework reveals a feasible path for scaling MLWP models to much higher spatial resolutions and finer timescales by performing the bulk of computation in a compressed latent representation.

\paragraph{Limitations and future work.}  
The current LaDCast model is constrained by the reconstruction error of its deep-compression autoencoder, which limits forecasting accuracy. Scaling encoder–decoder architectures to higher-resolution data has been shown to improve performance~\cite{han2024cra5}, and transformer-based autoencoders may better capture multimodal weather patterns~\cite{bodnar2024aurora}. Additionally, LaDCast is trained on reanalysis datasets, which are unavailable for real-time forecasting. We are exploring data assimilation techniques and multi-encoder frameworks~\cite{chen2024deep, huang2024diffda} to incorporate multi-modal real-time and near-real-time observations. Finally, the model’s spatial resolution limits analyses such as cyclone‐intensity estimation. Improved reconstruction metrics and physical‐consistency losses~\cite{jacobsen2023cocogenphysicallyconsistentconditionedscorebased} might help preserve high‐frequency details.

\begin{ack}
This work was supported by Los Alamos National Laboratory under the project “Algorithm/Software/Hardware Co-design for High Energy Density applications” at the University of Michigan. We also thank the C3S and WeatherBench2 teams for making the data publicly available.
\end{ack}

\clearpage
{
\small
\bibliographystyle{plain}
\bibliography{ref}
}


\appendix

\section{Broader impact.}  The adoption of a latent diffusion model (LDM) for global weather forecasting has the potential to democratize access to high-quality probabilistic forecasts by dramatically lowering the barriers imposed by storage and computational demands. By operating in a compact latent space, LaDCast reduces data storage and compute requirements by orders of magnitude compared to traditional NWP systems, enabling smaller institutions and even community-driven initiatives to run ensemble forecasting pipelines in near-real time. This increased accessibility can spur innovation in localized forecasting applications, enhance early-warning capabilities for extreme events, and support more equitable disaster preparedness efforts worldwide. In turn, LDM-based frameworks like LaDCast can broaden participation in weather and climate resilience, ensuring that the benefits of advanced forecasting reach underserved populations and contribute to global societal welfare. However, probabilistic forecasts may be misinterpreted by non-expert users, potentially leading to overconfidence in unlikely scenarios or under-preparation for high-impact, low-probability events.

\section{Diffusion model specification\label{app:diffusion}}


To generate samples conditioning on the input, we utilize the second-order Heun's solver in EDM~\cite{karras2022elucidating} to solve the probability-flow ODE Eq.~\ref{eq:pf_ode}. The solver requires \(2N-1\) steps with the last step using the first-order Euler method. The EDM noise schedule is defined as:
\[
  \sigma_t = \sigma_{\max}^\frac{1}{\rho} + \frac{t}{N-1} \left(\sigma_{\min}^\frac{1}{\rho} - \sigma_{\max}^\frac{1}{\rho}\right)^\rho
  \quad\text{for } t=0,1,\ldots,N-1,
\]
We utilize the default hyperparameters of EDM, with \(\sigma_{\max} = 80\), \(\sigma_{\min} = 0.002\) and \(\rho = 7\). The noise schedule is designed to be more aggressive at the beginning and less aggressive at the end. The number of reverse sampling steps is selected to be 20.

\section{Metrics\label{app:metric}}

The latitude weight, \(w(i)\), is defined as:
\begin{equation}
  w(i) = \frac{cos(\theta_i)}{\frac{1}{N}\sum_{i=1}^N cos(\theta_i)},
\end{equation}
where \(\theta_i\) is the latitude of the \(i\)-th grid point and \(N\) is the number of grid points. The latitude weight is equivalent to the area-based latitude weight in WB2 when the grid is uniform.

\paragraph{Latitude-weighted RMSE.} The latitude-weighted RMSE is defined as:
\begin{equation}
  \text{RMSE} = \sqrt{\frac{1}{N}\sum_{i=1}^N w(i) \left(\bm{y}_i - \bm{y}_{\text{pred},i}\right)^2},
\end{equation}

\paragraph{Continuous ranked probability score (CRPS).} The CRPS~\cite{gneiting2007strictly} is defined as: \(\mathbb{E}|X - Y| - \frac{1}{2}\mathbb{E}|X - X'|\), where \(X\) and \(Y\) are the predicted and observed values, respectively, and \(X'\) is a random sample from the predicted distribution. We follow the unbiased definition of CRPS~\cite{zamo2018estimation} that is also used in WB2:
\begin{equation}
  \text{CRPS} = w(i) \left( \frac{1}{N}\sum_{k=1}^N \left|x_{i,n}^k - y_{i,n}\right| -\frac{1}{2N(N-1)}\sum_{k=1}^N \sum_{k'=1}^N\left|x_{i,n}^k - x_{i,n}^{k'}\right|\right),
\quad \text{for } n=1,\ldots,N,
\end{equation}

\section{Training\label{app:train}}

\begin{table}[t]
  \centering
  \caption{EMA hyperparameters used during training}
  \label{tab:ema_hyperparams}
  \begin{tabular}{@{}lcl@{}}
    \toprule
    Parameter                           & Value    & Description                                              \\
    \midrule
    \texttt{ema\_max\_decay}            & 0.9999   & Target EMA decay rate $\alpha$                          \\
    \texttt{ema\_inv\_gamma}            & 1.0      & Inverse gamma for decay warmup schedule                 \\
    \texttt{ema\_power}                 & 0.6666667 & Power exponent in the warmup schedule                   \\
    \texttt{ema\_update\_after\_step}   & 1000     & Number of steps before starting EMA updates             \\
    \bottomrule
  \end{tabular}
\end{table}

All models are trained using the EMA hyperparameters shown in Table~\ref{tab:ema_hyperparams}. We use the final EMA weights for inference. Training was performed with the Accelerate library \cite{accelerate}, with all computations in bfloat16 precision.

\subsection{Data and preprocessing\label{app:data}}

We normalize all variables using the mean and standard deviation computed on the 1979–2017 training set, and apply these parameters uniformly across experiments. The \(1.5^\circ\) grid \((121 \times 240)\) is adjusted to an even latitude dimension by cropping the south pole, following common practice~\cite{bi2022pangu,bodnar2024aurora}. After normalization, NaN values over land in the sea surface temperature field are replaced with –2. The hourly dataset at \(1.5^\circ\) resolution (1979–2022) takes 3.0 TB to store, so we split it into monthly archives and load them via the Hugging Face Datasets library \cite{lhoest-etal-2021-datasets}, leveraging its chunk-shuffling feature to randomize samples during training. Once the DC-AE is trained, we normalize its latent outputs again, standardizing each latent channel independently using the original training-set (1979-2017) for training the downstream autoregressive model.

\subsection{Deep Compression AutoEncoder\label{app:train_auto}}

\begin{table}[t]
  \centering
  \caption{Optimizer and LR schedule hyperparameters for pretraining and finetuning the DC-AE}
  \label{tab:train_hyperparams}
  \begin{tabular}{@{}lcc@{}}
    \toprule
    Parameter             & Pretraining          & Finetuning           \\
    \midrule
    Optimizer             & AdamW                & AdamW                \\
    Betas                 & (0.9, 0.999)         & (0.9, 0.999)         \\
    $\epsilon$            & $1\times10^{-8}$     & $1\times10^{-8}$     \\
    Learning Rate         & $1\times10^{-4}$     & $1\times10^{-5}$     \\
    Weight Decay          & $1\times10^{-2}$     & $1\times10^{-2}$     \\
    LR Scheduler          & Cosine               & Cosine               \\
    Warmup Steps          & 1000                 & 1000                 \\
    \bottomrule
  \end{tabular}
\end{table}

The training hyperparameters for the DC-AE are given in Table \ref{tab:train_hyperparams}. Pretraining requires 141 hours on eight H100 GPUs, while fine-tuning takes 54 hours on four H100 GPUs. We train for 30 epochs during pretraining and 10 epochs during fine-tuning. To increase throughput, we apply a simple data‐augmentation procedure in both phases: each step we randomly select a pair of coordinates and shift them to the top-left corner of the grid. We perform two such augmentation steps per normal training step; these do not count as additional iterations, but gradients are backpropagated normally for each.

\begin{table}[t]
  \centering
  \caption{DC-AE architecture. Decoder mirrors the encoder.}
  \label{tab:dcae_arch}
  \begin{tabular}{@{}lcccc@{}}
    \toprule
    Stage               & Block Type          & Out Channels & Layers & QKV Multiscales \\
    \midrule
    Encoder 1           & ResBlock            & 252          & 4         & –               \\
    Encoder 2           & ResBlock            & 504          & 4         & –               \\
    Encoder 3           & EfficientViTBlock   & 504          & 4         & 5             \\
    Encoder 4           & EfficientViTBlock   & 1008         & 4         & 5             \\
    Latent projection & SphericalCNN & 84  & – & – \\
    \midrule
    Decoder (mirrors)   & —                   & —            & —         & —               \\
    \bottomrule
  \end{tabular}
\end{table}

\begin{figure}[htb]
  \centering
  \includegraphics[width=1.0\textwidth]{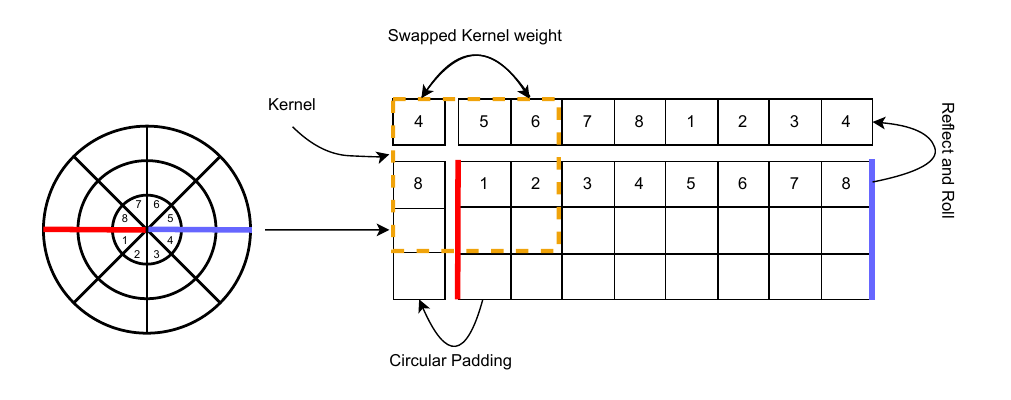}
  \caption{An illustration of the spherical convolutional kernel. Kernels' weights at the corresponding rows at the top and bottom will be swapped to address spherical nature of the data.}
    \label{fig:pole}
\end{figure}

The DC-AE architecture is summarized in Table~\ref{tab:dcae_arch}. It takes 89 input channels, including 5 static features, and produces 89 output channels. The compressed latent representation contains 84 channels, corresponding to the number of dynamic input variables. To account for the data’s spherical geometry, all convolutions are replaced with our proposed spherical CNN block (see Figure~\ref{fig:pole}). The model is trained using a batch size of 4.

\subsection{LaDCast\label{app:train_ladcast}}

\begin{table}[t]
  \centering
  \caption{Architecture hyperparameters for LaDCast (1.6B and 375M).}
  \label{tab:ar_model_arch}
  \begin{tabular}{@{}lcc@{}}
    \toprule
                           & \textbf{1.6B} & \textbf{375M} \\
    \midrule
     Layers               & 5             & 2             \\
     Preprocess Layers       & 3             & 1             \\
     Single‐stream Layers & 10            & 4             \\
    Attention Heads         & 16            & 12            \\
    Head Dim                & 128           & 128           \\
    Model Dim               & 2048          & 1536          \\
    FFN Dim                 & 8192          & 6144          \\
    Rope Axes (d\textsubscript{t}, d\textsubscript{h}, d\textsubscript{w}) & (16, 56, 56)  & (16, 56, 56)  \\
    \bottomrule
  \end{tabular}
\end{table}

The architectures of the 1.6B and 375M autoregressive models are summarized in Table~\ref{tab:ar_model_arch}. We use the same optimizer and learning‐rate schedule as in the DC-AE pretraining. The 1.6B variant is employed for our main results, while the 375M variant is used in ablation studies. Both models are trained with a batch size of four and optimized with the default weighted loss from EDM \cite{karras2022elucidating}:
\[
  \text{loss} = \frac{\sigma^2 + \sigma_{\text{data}}^2}{\sigma \cdot \sigma_{\text{data}}} \cdot \left(\bm{y} - \bm{y}_{\text{pred}}\right)^2,
\]
where \(\sigma\) is the noise level and \(\sigma_{\text{data}}\) is selected to be \(0.5\). We further added a latitude-weighted term for weighting the denoising loss. During training, \(\sigma\) is sampled from the log-normal distribution proposed in EDM~\cite{karras2022elucidating} (using default parameters), with the sampler seeded by the current training step for reproducibility. On 8 H100 GPUs, the 1.6B model trains in about 16 hours (with gradient checkpointing enabled), while the 375M model completes in 4 hours.

\section{Supplementary information\label{app:info}}

\begin{table}[t]
  \caption{Overview of model variables: atmospheric, single‑level, static and clock features.}
  \label{tab:variables}
  \centering
  \begin{tabular}{llccl}
    \toprule
    \textbf{Type}
      & \textbf{Variable name}
      & \textbf{Short name}
      & \makecell[c]{\textbf{ECMWF}\\\textbf{Param ID}}
      & \makecell[l]{\textbf{Role}}\\
    \midrule
    Atmospheric & Geopotential                 & z    & 129 & Input/Predicted       \\
    Atmospheric & Specific humidity            & q    & 133 & Input/Predicted       \\
    Atmospheric & Temperature                  & t    & 130 & Input/Predicted       \\
    Atmospheric & U component of wind          & u    & 131 & Input/Predicted       \\
    Atmospheric & V component of wind          & v    & 132 & Input/Predicted       \\
    Atmospheric & Vertical velocity            & w    & 135 & Input/Predicted       \\
    Single      & 10 metre u wind component    & 10u  & 165 & Input/Predicted       \\
    Single      & 10 metre v wind component    & 10v  & 166 & Input/Predicted       \\
    Single      & 2 metre temperature          & 2t   & 167 & Input/Predicted       \\
    Single      & Mean sea level pressure      & msl  & 151 & Input/Predicted       \\
    Single      & Sea Surface Temperature      & sst  &  34 & Input/Predicted       \\
    Single      & Total precipitation (6h, derived)         & tp   & 228 & Input/Predicted       \\
    \midrule
    Static      & Land–sea mask                & lsm  & 172 & Input                 \\
    Static      & Standard deviation of orography                     & sdor  & 160 & Input                 \\
    Static      & Angle of sub-gridscale orography                    & isor  & 162 & Input                 \\
    Static   & Anisotropy of sub-gridscale orography & anor & 161 & Input \\
    Static   & Slope of sub-gridscale orography & slor & 163 & Input    \\
    Clock       & Elapsed year progress        & n/a  & n/a & Input                 \\
    \bottomrule
  \end{tabular}
\end{table}

\subsection{Deep compression Autoencoder architecture\label{app:autoencoder}}

Below we provide a sketch proof of convoluting with static features at the first layer is equivalent to patchifying the variables with some embedding based on the static features. We consider a sequence of $T$ high‐dimensional fields \(\{x^t\}_{t=1}^T \subset \mathcal{X}\), \(x^t \in \mathbb{R}^{C \times H \times W}
\) where we decompose the channel dimension $C = C_s + C_v$ into
\begin{equation}
x^t
= \bigl[x_s,\;x_v^t\bigr]
\,,
\quad
x_s \in \mathbb{R}^{C_s\times H\times W}
\quad
\text{(static)},
\quad
x_v^t \in \mathbb{R}^{C_v\times H\times W}
\quad
\text{(variable)}.
\end{equation}
With the convolution layer parameterized by \(W \in \mathbb{R}^{C'\times C \times K \times K},\, b \in \mathbb{R}^{C'}\), we split the kernel as
\begin{equation}
W = \bigl[\,W_s \;\big|\; W_v\bigr]
\,,\quad
W_s \in \mathbb{R}^{C'\times C_s\times K\times K}
\,,\;
W_v \in \mathbb{R}^{C'\times C_v\times K\times K}.
\end{equation}
Then for each $t$,
\begin{equation}
y^t
= W * x^t + b = W_s * x_s \;+\; W_v * x_v^t \;+\; b
.
\end{equation}
Since $x_s$ is fixed, define the static embedding, \(E\;=\;W_s * x_s + b
\;\in\;\mathbb{R}^{C'\times H'\times W'}
\), which does not depend on $t$.  Hence
\begin{equation}
y^t
= \underbrace{\bigl(W_v * x_v^t\bigr)}_{\text{patchify}}\;+\;E.
\end{equation}

\subsection{LadCast embedding\label{app:arch_ladcast}}

\paragraph{Periodic elapsed year embedding.} 
To effectively model seasonal patterns and capture the inherently cyclic nature of annual time, we represent each timestamp by its fraction of the year elapsed and project this scalar onto a fixed set of sinusoidal basis functions. Such a periodic encoding ensures continuity at year boundaries. Let \(D\) be the embedding dimensionality and \(K = \frac{D}{2}\). We can then define the phase based on the elapsed year:
\[
\psi^i = 2 \pi \,\frac{t^i - t^0}{T^i}
\quad\in[0,2\pi],
\]
where \(t^i \text{ is the raw timestamp, }
t^{0} \text{ is the start of the year for } t^i,\text{ and }
T^i \text{ is the length of that year.}\) For each frequency index \(k=1,2,\ldots,K\), define:
\[
\alpha_k =
\exp\Bigl(-\,\frac{\ln P}{K}\,(k-1)\Bigr),
\quad P = 10000,
\]
and form the embedding components: \(e^i_{2k-1} = \alpha_k \,\sin\bigl(k\,\psi^i\bigr),\, e^i_{2k}   = \alpha_k \,\cos\bigl(k\,\psi^i\bigr).\)
Finally, we concatenate these into the full \(D\)-dimensional vector
\[
\mathbf{e}^i 
= \bigl[e^i_{1},\,e^i_{2},\,\dots,\,e^i_{2K-1},\,e^i_{2K}\bigr]^\top
\;\in\;\mathbb{R}^D.
\]

\paragraph{Geometric Rotary Position Embedding}

The original Rotary Position Embedding (RoPE)~\cite{su2024roformer} is designed for 1D data. We adapt it to create GeoRoPE, which accommodates the periodicity in the longitude dimension and the atmospheric circulation patterns in the latitude dimension. 

We begin by reviewing the 1D RoPE. Let $d$ be the model hidden dimension (assumed even). RoPE applies a complex rotation $e^{i\theta}$ to key and query vectors, resulting in an attention matrix expressible as:
\[
  A_{(n,m)} = \text{Re}\big[\bm{q}_n\bm{k}_m^* e^{i(n-m)\theta}\big],
\]
where $\bm{q}_n$ and $\bm{k}_m$ are the $n$-th and $m$-th query and key vectors, respectively. Here, $\text{Re}$ denotes the real part and $*$ denotes the complex conjugate. This attention operation effectively encodes the relative positional information $(n-m)$.

To extend RoPE to geographical data, we apply the 1D operation for each dimension with appropriate modifications. For 2D/3D data, we adopt a higher frequency scale~\cite{heo2024rotary}, setting $\theta_j = 256^{-2j/d}$ for the $j$-th frequency dimension where $j \in \{0, 1, \ldots, d/2-1\}$. 

For geographical data, we encode positions using the center coordinates $(p_{\text{lon}}, p_{\text{lat}})$ of each patch. For longitude, conventional encoding would use indices $\{0,1,\ldots,W-1\}$ for width $W$. Instead, we map longitude to $p_{\text{lon}} \in [0,2\pi)$ to capture its inherent periodicity. For latitude, which lacks the same periodicity, we adopt a mapping based on atmospheric circulation patterns. Zonal winds (parallel to latitude) exhibit alternating circulation bands~\cite{schneider2006general}, motivating our mapping to $p_{\text{lat}} \in [-1.5\pi,1.5\pi]$ for accounting the three alternating circulation at each hemisphere. The attention computation between patches at positions $(p_{\text{lon}}^n, p_{\text{lat}}^n)$ and $(p_{\text{lon}}^m, p_{\text{lat}}^m)$ becomes:
\[
A_{(n,m)} = \text{Re}\left[\bm{q}_n\bm{k}_m^* e^{i(p_{\text{lon}}^n-p_{\text{lon}}^m)\theta_{\text{lon}}} e^{i(p_{\text{lat}}^n-p_{\text{lat}}^m)\theta_{\text{lat}}}\right]
\]

To extend to the temporal dimension, we follow the basic extension and index from \(-1\) (since we set input to be one) to \(T_\text{output}\), which is the length of the output sequence. For \(d=128\) (Table~\ref{tab:ar_model_arch}), we partition the temporal, latitude and longitude parts as 16, 56 and 56, respectively.

\section{Supplementary results\label{app:results}}

\subsection{Autoencoder\label{app:ae}}

\begin{longtable}{lrrrrr}
\caption{Validation RMSE by variable and year} \label{tab:rmse_by_year} \\
\toprule
Short name & 2018 & 2019 & 2020 & 2021 & 2022 \\
\midrule
\endfirsthead
\caption[]{Validation RMSE by variable and year} \\
\toprule
Short name & 2018 & 2019 & 2020 & 2021 & 2022 \\
\midrule
\endhead
\midrule
\multicolumn{6}{r}{Continued on next page} \\
\midrule
\endfoot
\bottomrule
\endlastfoot
10u & 5.478e-01 & 5.520e-01 & 5.485e-01 & 5.461e-01 & 5.476e-01 \\
10v & 5.368e-01 & 5.414e-01 & 5.369e-01 & 5.348e-01 & 5.357e-01 \\
2t & 7.819e-01 & 7.793e-01 & 7.836e-01 & 7.851e-01 & 8.149e-01 \\
anor & 2.130e-03 & 2.134e-03 & 2.160e-03 & 2.141e-03 & 2.153e-03 \\
isor & 5.153e-03 & 5.171e-03 & 5.233e-03 & 5.195e-03 & 5.230e-03 \\
sdor & 4.687e-01 & 4.709e-01 & 4.746e-01 & 4.716e-01 & 4.732e-01 \\
slor & 7.704e-05 & 7.736e-05 & 7.802e-05 & 7.767e-05 & 7.786e-05 \\
sst & 6.492e-01 & 6.249e-01 & 6.459e-01 & 6.253e-01 & 6.863e-01 \\
$l_2$-norm loss & 1.508e-01 & 1.517e-01 & 1.512e-01 & 1.519e-01 & 1.528e-01 \\
lsm & 3.032e-03 & 3.058e-03 & 3.065e-03 & 3.053e-03 & 3.036e-03 \\
msl & 3.027e+01 & 3.073e+01 & 3.255e+01 & 3.153e+01 & 3.316e+01 \\
tp-6h & 4.216e-04 & 4.234e-04 & 4.220e-04 & 4.185e-04 & 4.175e-04 \\
q-50 & 2.440e-08 & 2.950e-08 & 2.614e-08 & 2.708e-08 & 2.865e-08 \\
q-100 & 1.141e-07 & 1.202e-07 & 1.106e-07 & 1.141e-07 & 1.085e-07 \\
q-150 & 8.539e-07 & 9.034e-07 & 9.026e-07 & 8.605e-07 & 8.521e-07 \\
q-200 & 5.123e-06 & 5.389e-06 & 5.441e-06 & 5.143e-06 & 5.052e-06 \\
q-250 & 1.747e-05 & 1.820e-05 & 1.840e-05 & 1.762e-05 & 1.742e-05 \\
q-300 & 3.940e-05 & 4.056e-05 & 4.131e-05 & 4.000e-05 & 3.983e-05 \\
q-400 & 1.231e-04 & 1.258e-04 & 1.283e-04 & 1.249e-04 & 1.252e-04 \\
q-500 & 2.508e-04 & 2.551e-04 & 2.586e-04 & 2.530e-04 & 2.531e-04 \\
q-600 & 3.900e-04 & 3.932e-04 & 3.977e-04 & 3.908e-04 & 3.914e-04 \\
q-700 & 5.608e-04 & 5.683e-04 & 5.702e-04 & 5.618e-04 & 5.644e-04 \\
q-850 & 7.071e-04 & 7.235e-04 & 7.212e-04 & 7.054e-04 & 7.140e-04 \\
q-925 & 5.769e-04 & 5.948e-04 & 5.905e-04 & 5.837e-04 & 5.901e-04 \\
q-1000 & 4.458e-04 & 4.528e-04 & 4.588e-04 & 4.587e-04 & 4.736e-04 \\
t-50 & 6.542e-01 & 6.854e-01 & 6.761e-01 & 6.494e-01 & 7.175e-01 \\
t-100 & 6.500e-01 & 6.644e-01 & 7.005e-01 & 6.854e-01 & 7.026e-01 \\
t-150 & 4.689e-01 & 4.755e-01 & 4.907e-01 & 4.852e-01 & 4.908e-01 \\
t-200 & 4.310e-01 & 4.352e-01 & 4.460e-01 & 4.397e-01 & 4.485e-01 \\
t-250 & 4.228e-01 & 4.312e-01 & 4.336e-01 & 4.290e-01 & 4.405e-01 \\
t-300 & 4.244e-01 & 4.314e-01 & 4.352e-01 & 4.331e-01 & 4.422e-01 \\
t-400 & 4.690e-01 & 4.795e-01 & 4.832e-01 & 4.816e-01 & 4.868e-01 \\
t-500 & 4.779e-01 & 4.864e-01 & 4.909e-01 & 4.899e-01 & 4.950e-01 \\
t-600 & 4.960e-01 & 5.013e-01 & 5.055e-01 & 5.034e-01 & 5.093e-01 \\
t-700 & 5.251e-01 & 5.288e-01 & 5.338e-01 & 5.317e-01 & 5.375e-01 \\
t-850 & 6.503e-01 & 6.591e-01 & 6.641e-01 & 6.602e-01 & 6.669e-01 \\
t-925 & 5.968e-01 & 6.095e-01 & 6.063e-01 & 6.063e-01 & 6.262e-01 \\
t-1000 & 6.449e-01 & 6.485e-01 & 6.554e-01 & 6.615e-01 & 6.986e-01 \\
u-50 & 1.184e+00 & 1.265e+00 & 1.219e+00 & 1.225e+00 & 1.231e+00 \\
u-100 & 1.366e+00 & 1.398e+00 & 1.398e+00 & 1.394e+00 & 1.403e+00 \\
u-150 & 1.548e+00 & 1.573e+00 & 1.574e+00 & 1.568e+00 & 1.581e+00 \\
u-200 & 1.615e+00 & 1.621e+00 & 1.617e+00 & 1.621e+00 & 1.634e+00 \\
u-250 & 1.603e+00 & 1.605e+00 & 1.605e+00 & 1.606e+00 & 1.621e+00 \\
u-300 & 1.662e+00 & 1.663e+00 & 1.670e+00 & 1.667e+00 & 1.681e+00 \\
u-400 & 1.561e+00 & 1.562e+00 & 1.578e+00 & 1.571e+00 & 1.585e+00 \\
u-500 & 1.394e+00 & 1.395e+00 & 1.409e+00 & 1.404e+00 & 1.411e+00 \\
u-600 & 1.281e+00 & 1.283e+00 & 1.295e+00 & 1.290e+00 & 1.294e+00 \\
u-700 & 1.241e+00 & 1.244e+00 & 1.254e+00 & 1.248e+00 & 1.252e+00 \\
u-850 & 1.015e+00 & 1.022e+00 & 1.023e+00 & 1.017e+00 & 1.019e+00 \\
u-925 & 7.727e-01 & 7.774e-01 & 7.751e-01 & 7.720e-01 & 7.723e-01 \\
u-1000 & 6.035e-01 & 6.084e-01 & 6.047e-01 & 6.024e-01 & 6.036e-01 \\
v-50 & 9.989e-01 & 1.044e+00 & 1.011e+00 & 1.005e+00 & 1.010e+00 \\
v-100 & 1.173e+00 & 1.201e+00 & 1.198e+00 & 1.190e+00 & 1.197e+00 \\
v-150 & 1.331e+00 & 1.352e+00 & 1.351e+00 & 1.343e+00 & 1.356e+00 \\
v-200 & 1.445e+00 & 1.456e+00 & 1.447e+00 & 1.447e+00 & 1.457e+00 \\
v-250 & 1.523e+00 & 1.530e+00 & 1.525e+00 & 1.523e+00 & 1.537e+00 \\
v-300 & 1.569e+00 & 1.577e+00 & 1.578e+00 & 1.572e+00 & 1.585e+00 \\
v-400 & 1.482e+00 & 1.490e+00 & 1.498e+00 & 1.491e+00 & 1.501e+00 \\
v-500 & 1.317e+00 & 1.325e+00 & 1.334e+00 & 1.329e+00 & 1.333e+00 \\
v-600 & 1.210e+00 & 1.218e+00 & 1.227e+00 & 1.220e+00 & 1.226e+00 \\
v-700 & 1.171e+00 & 1.180e+00 & 1.186e+00 & 1.180e+00 & 1.184e+00 \\
v-850 & 9.880e-01 & 9.987e-01 & 9.958e-01 & 9.903e-01 & 9.942e-01 \\
v-925 & 7.534e-01 & 7.609e-01 & 7.562e-01 & 7.521e-01 & 7.538e-01 \\
v-1000 & 5.920e-01 & 5.977e-01 & 5.929e-01 & 5.908e-01 & 5.916e-01 \\
w-50 & 7.062e-03 & 6.963e-03 & 6.958e-03 & 6.915e-03 & 7.064e-03 \\
w-100 & 1.425e-02 & 1.453e-02 & 1.433e-02 & 1.413e-02 & 1.421e-02 \\
w-150 & 2.551e-02 & 2.590e-02 & 2.570e-02 & 2.531e-02 & 2.552e-02 \\
w-200 & 3.422e-02 & 3.467e-02 & 3.436e-02 & 3.417e-02 & 3.440e-02 \\
w-250 & 4.242e-02 & 4.273e-02 & 4.250e-02 & 4.249e-02 & 4.267e-02 \\
w-300 & 5.241e-02 & 5.270e-02 & 5.241e-02 & 5.235e-02 & 5.257e-02 \\
w-400 & 6.726e-02 & 6.750e-02 & 6.700e-02 & 6.676e-02 & 6.714e-02 \\
w-500 & 6.963e-02 & 6.984e-02 & 6.947e-02 & 6.928e-02 & 6.974e-02 \\
w-600 & 7.238e-02 & 7.291e-02 & 7.252e-02 & 7.231e-02 & 7.255e-02 \\
w-700 & 7.715e-02 & 7.783e-02 & 7.705e-02 & 7.693e-02 & 7.711e-02 \\
w-850 & 7.175e-02 & 7.227e-02 & 7.185e-02 & 7.147e-02 & 7.174e-02 \\
w-925 & 5.299e-02 & 5.335e-02 & 5.313e-02 & 5.283e-02 & 5.311e-02 \\
w-1000 & 1.796e-02 & 1.807e-02 & 1.801e-02 & 1.794e-02 & 1.817e-02 \\
z-50 & 7.488e+01 & 8.578e+01 & 1.198e+02 & 9.345e+01 & 9.146e+01 \\
z-100 & 5.103e+01 & 6.582e+01 & 7.352e+01 & 6.912e+01 & 6.831e+01 \\
z-150 & 4.055e+01 & 5.183e+01 & 5.586e+01 & 5.375e+01 & 6.015e+01 \\
z-200 & 3.696e+01 & 4.531e+01 & 4.652e+01 & 4.780e+01 & 5.628e+01 \\
z-250 & 3.520e+01 & 4.558e+01 & 4.619e+01 & 4.914e+01 & 5.846e+01 \\
z-300 & 3.466e+01 & 4.283e+01 & 4.240e+01 & 4.533e+01 & 5.379e+01 \\
z-400 & 3.085e+01 & 3.698e+01 & 3.689e+01 & 3.849e+01 & 4.563e+01 \\
z-500 & 2.896e+01 & 3.419e+01 & 3.498e+01 & 3.619e+01 & 4.259e+01 \\
z-600 & 2.724e+01 & 3.145e+01 & 3.273e+01 & 3.394e+01 & 3.938e+01 \\
z-700 & 2.443e+01 & 2.702e+01 & 2.896e+01 & 2.955e+01 & 3.391e+01 \\
z-850 & 2.143e+01 & 2.242e+01 & 2.416e+01 & 2.417e+01 & 2.721e+01 \\
z-925 & 2.165e+01 & 2.240e+01 & 2.413e+01 & 2.382e+01 & 2.619e+01 \\
z-1000 & 2.378e+01 & 2.426e+01 & 2.579e+01 & 2.500e+01 & 2.648e+01 \\
\end{longtable}

\subsection{LaDCast}

\begin{figure}[htb]
  \centering
  \includegraphics[width=1.0\textwidth]{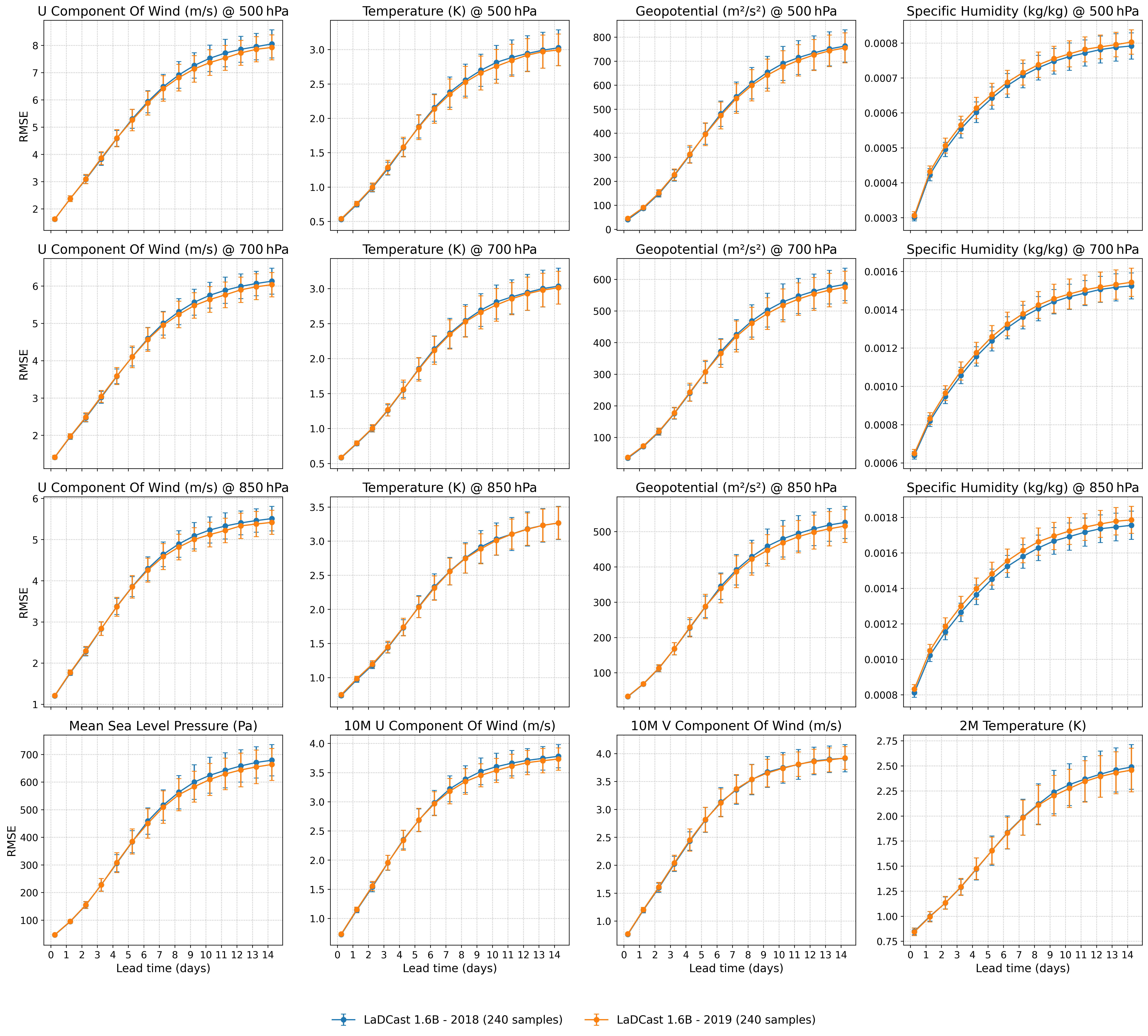}
  \caption{Latitude‑weighted RMSE of LaDCast for 2018 and 2019. Error bars indicate the standard deviation computed from 240 samples.}
  \label{fig:ladcast_2018_2019}
\end{figure}

\begin{figure}[htb]
  \centering
  \includegraphics[width=1.0\textwidth]{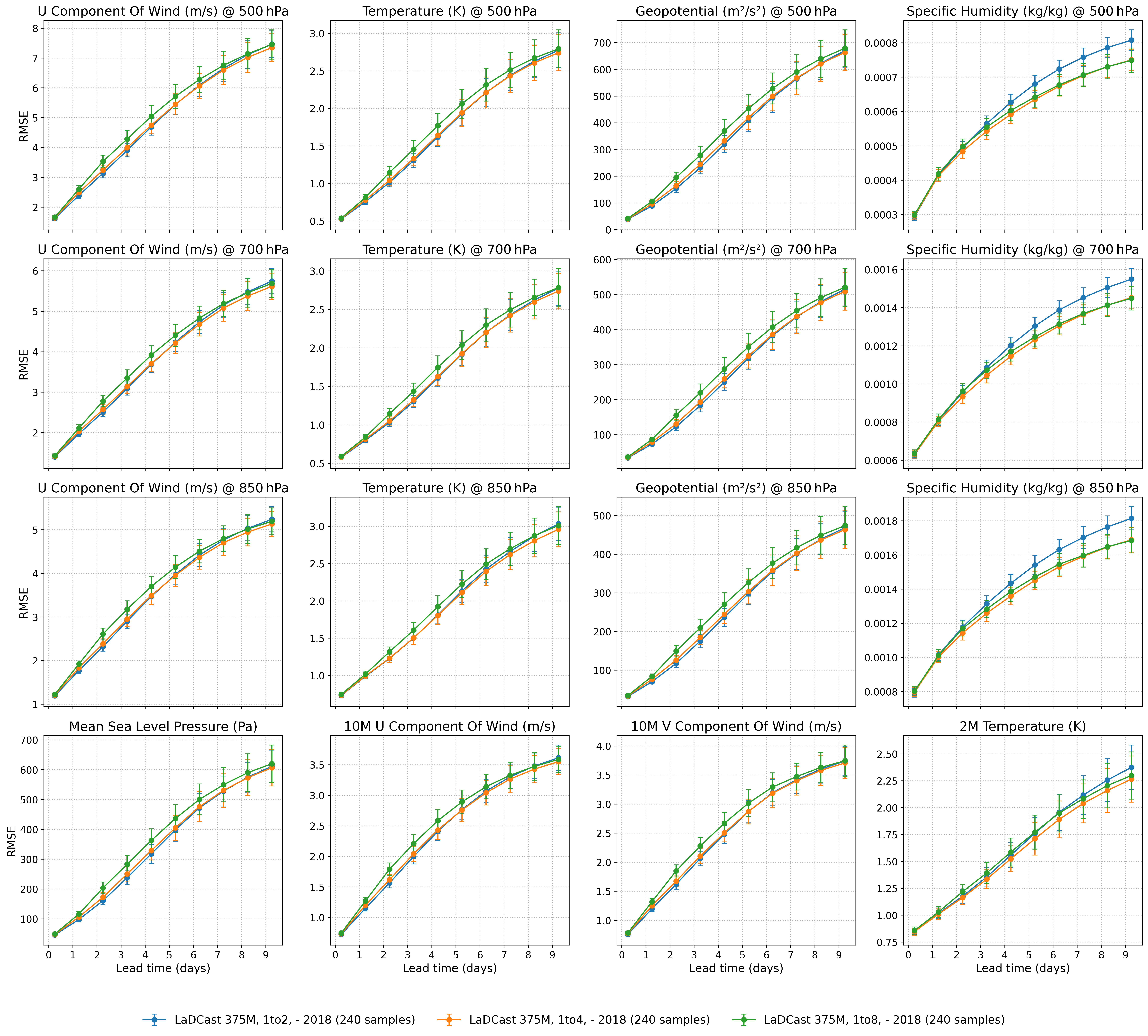}
  \caption{Ablation study of the effect of input‑sequence length.}
  \label{fig:ladcast_input_sequence}
\end{figure}

\begin{figure}[htb]
  \centering
  \includegraphics[width=1.0\textwidth]{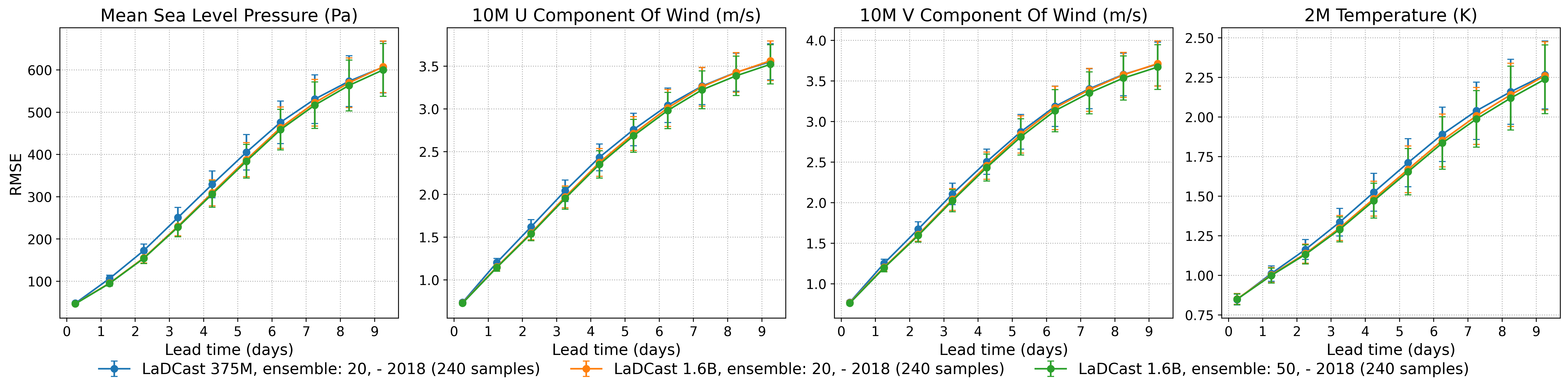}
  \caption{Ablation study on scaling and ensemble size.}
  \label{fig:ladcast_scaling}
\end{figure}

\begin{figure}[htb]
  \centering
  \includegraphics[width=1.0\textwidth]{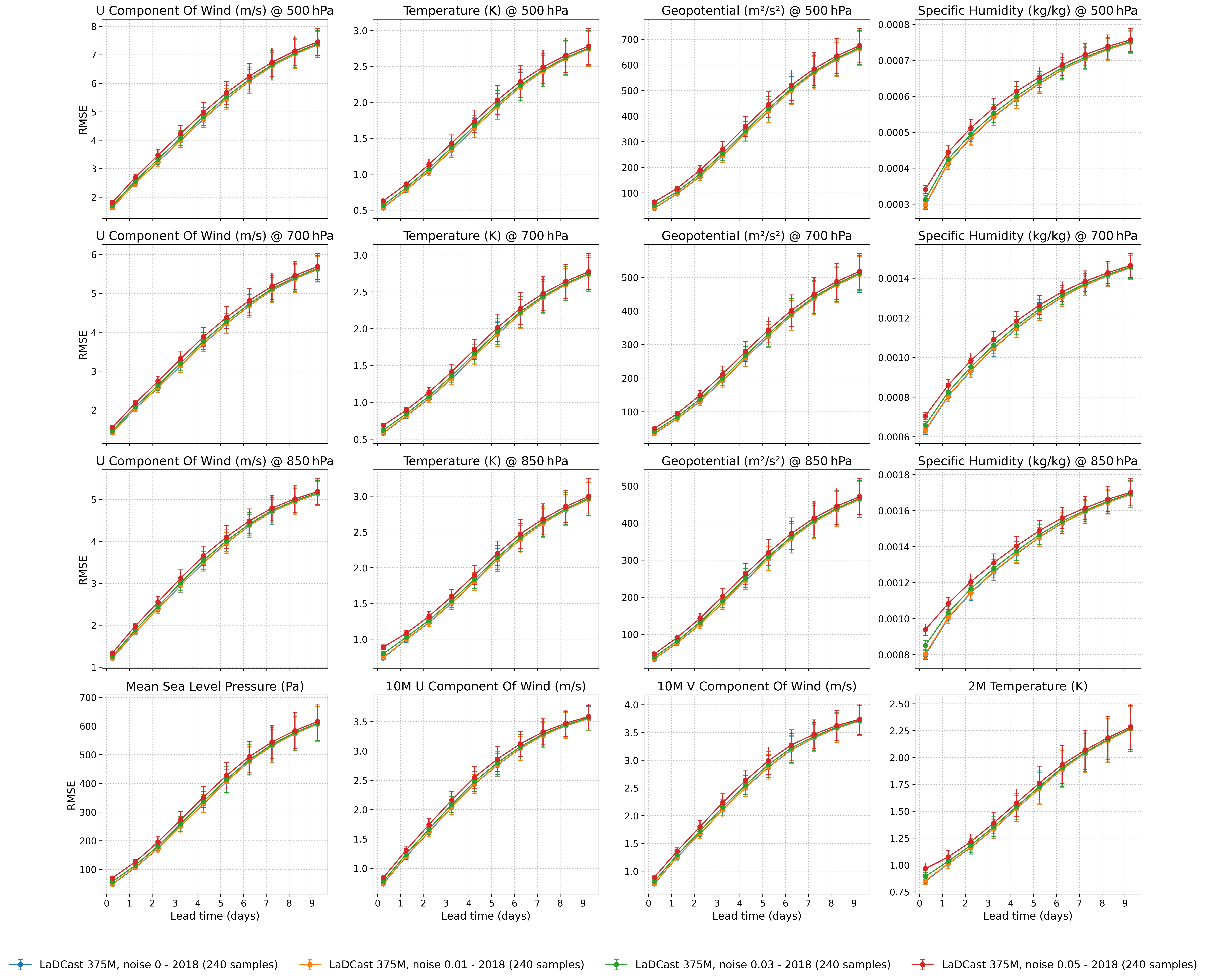}
  \caption{Ablation study of latent‑space perturbations.}
  \label{fig:ladcast_noise}
\end{figure}

\begin{figure}[htb]
  \centering
  \includegraphics[width=1.0\textwidth]{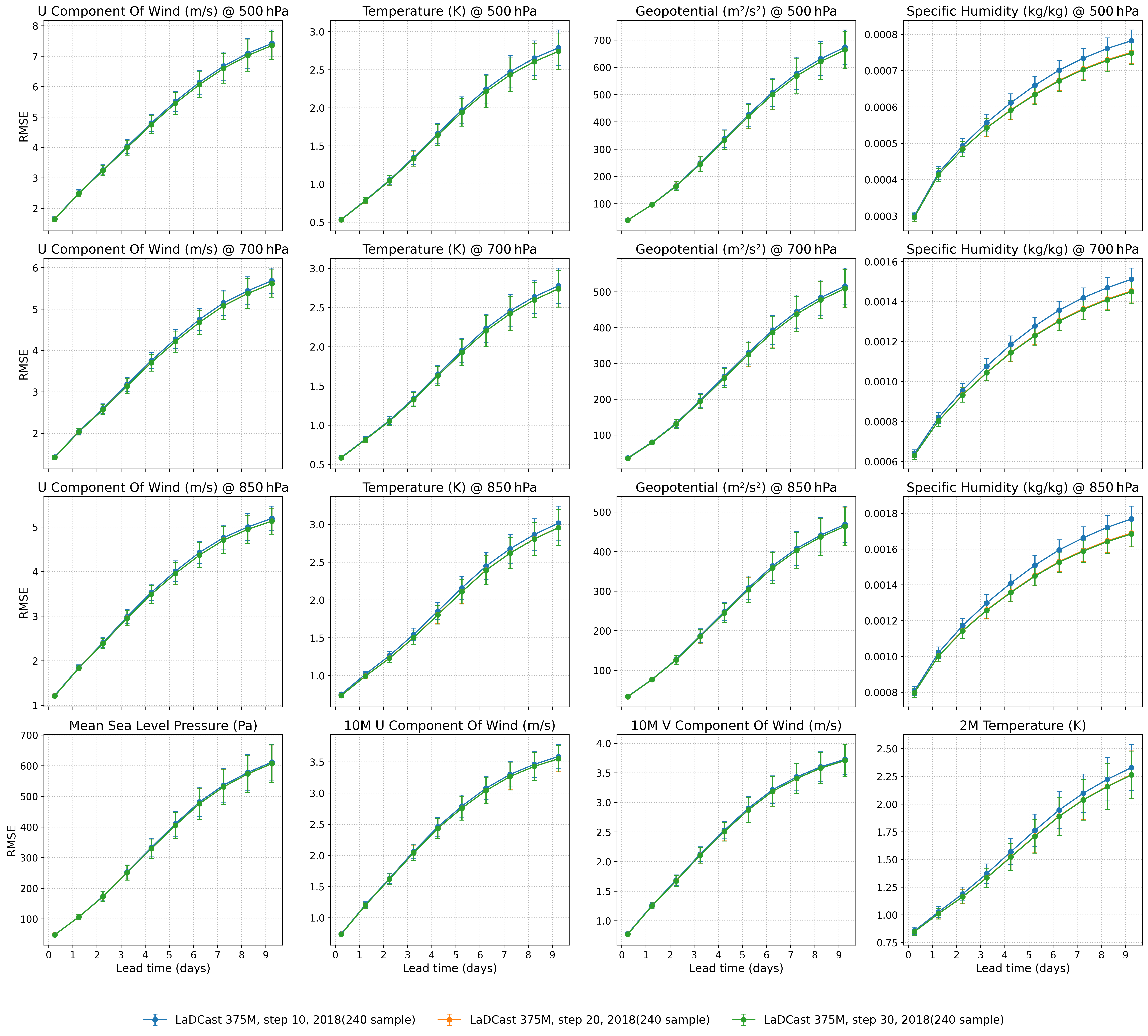}
  \caption{Ablation study on the number of reverse‑sampling steps.}
  \label{fig:ladcast_steps}
\end{figure}

We initialize the Gaussian noise prior for all results with a seed tied to each ensemble member’s index to ensure reproducibility. In the ablation studies, we scale RMSE error bars by the standard deviation of RMSE computed from 240 samples. For clarity, these error bars are omitted from the main text.

\section{Cyclone tracking \label{app:tracking}}

This appendix describes the implementation of a heuristic cyclone tracker based on the approach outlined in~\cite{bodnar2024aurora}. The tracker follows local minima in mean sea level pressure and utilizes geopotential height at 700 hPa when necessary. In practice, we only utilize geopotential for the hurricane Dorian, for the other two, we enforce mean sea level pressure for faster evaluation.

In this implementation, a point is classified as a local minimum if it has the lowest value within a small box centered around that point. The size of this validation box is determined by the inner degree parameter with a \(1.5^\circ\) buffer. The tracker searches for local minima within larger search boxes of varying sizes. If a detected minimum is located on the edge of the search box, it is considered invalid and that particular search is declared a failure.

The tracking algorithm operates as follows:
\begin{enumerate}
    \item Initialization: The tracker starts with a known initial position (latitude-longitude) of a tropical cyclone at time \(t_0\).
    \item Grid Alignment: All positional values are rounded to align with the model grid resolution for consistency.
    \item  Iterative Tracking: For each 6-hour time step: a. The current position is used as the center point for the search. b. If the current position is over water (determined by land-sea mask when enabled, however, we only enable this for tracking Dorian), the algorithm searches for local minima in mean sea level pressure. c. The search occurs in progressively smaller boxes [7,4,1] for Lorenzo, [7,5,1] for Kong-rey and [6,3,0] for Dorian, until a valid minimum is found. d. A minimum is only accepted if it represents a position change from the previous point. e. If no valid MSLP minimum is found or if the position doesn't change, the algorithm attempts to find a local minimum in Z700 using the same box sizes (we only enable this for tracking Dorian). f. If no valid minimum is found in either field, the position remains unchanged and appropriate warnings are issued. g. The position (either updated or unchanged) is recorded for the current time step.
\end{enumerate}

\section{License\label{app:license}}
\begin{itemize}
    \item \textbf{LaDCast}: LaDCast contains codes from the diffusers library~\cite{von-platen-etal-2022-diffusers} which is licensed under the Apache License 2.0.
    \item \textbf{ERA5}: The ERA5 dataset is licensed under the Copernicus License. Use of ERA5 data is free of charge, worldwide, non-exclusive, royalty-free and perpetual.

\end{itemize}

\end{document}